\documentclass[preprint,5p,times,twocolumn]{elsarticle}
\usepackage{graphicx}
\usepackage{amssymb}
\usepackage{multirow}
\usepackage{lineno}

\makeatletter
\def\ps@pprintTitle{%
	\let\@oddhead\@empty
	\let\@evenhead\@empty
	\def\@oddfoot{\reset@font\hfil\thepage\hfil}%
	\let\@evenfoot\@oddfoot}
\makeatother

\begin{document}

\begin{frontmatter}


\title{Histopathologic Image Processing: A Review}



\author[ETS,UEPG]{Jonathan de Matos}
\author[PUCPR]{Alceu de Souza Britto Junior}
\author[UFPR]{Luiz Eduardo S. Oliveira}
\author[ETS]{Alessandro Lameiras Koerich}

\address[ETS]{Ecole de Technologie Superiéure, Montreal, Canada}
\address[UEPG]{Universidade Estadual de Ponta Grossa, Ponta Grossa, Brasil}
\address[PUCPR]{Pontificia Universidade Catolica do Parana, Curitiba, Brasil}
\address[UFPR]{Universidade Federal do Parana, Curitiba, Brasil}

%%%%%%%%%%%%%%%%%%%%%%%%%%%%%%%%%%%%%%%%%%%%%%%%%%%%%%%%%%%%%%%%%%%%
\begin{abstract}
Histopathologic Images (HI) are the gold standard for evaluation of some tumors. However, the analysis of such images is challenging even for experienced pathologists, resulting in problems of inter and intra observer. Besides that, the analysis is time and resource consuming. One of the ways to accelerate such an analysis is by using Computer Aided Diagnosis systems. In this work we present a literature review about the computing techniques to process HI, including shallow and deep methods. We cover the most common tasks for processing HI such as segmentation, feature extraction, unsupervised learning and supervised learning. A dataset section show some datasets found during the literature review. We also bring a study case of  breast cancer classification using a mix of deep and shallow machine learning methods. The proposed method obtained an accuracy of 91\% in the best case, outperforming the compared baseline of the dataset.
\end{abstract}
%%%%%%%%%%%%%%%%%%%%%%%%%%%%%%%%%%%%%%%%%%%%%%%%%%%%%%%%%%%%%%%%%%%%

\end{frontmatter}
%%%%%%%%%%%%%%%%%%%%%%%%%%%%%%%%%%%%%%%%%%%%%%%%%%%%%%%%%%%%%%%%%%%%

%%\linenumbers

%%%%%%%%%%%%%%%%%%%%%%%%%%%%%%%%%%%%%%%%%%%%%%%%%%%%%%%%%%%%%%%%%%%%
\section{Introduction}
%%%%%%%%%%%%%%%%%%%%%%%%%%%%%%%%%%%%%%%%%%%%%%%%%%%%%%%%%%%%%%%%%%%%
Current hardware capabilities and computing technologies provide the ability of computing to solve problems in many fields. The medical field is a noble employ of technology as it can help to improve populations' health and quality of life. Medical diagnosis is a good example of the application of computing. One type of diagnosis is the one based on images, e.g. Magnetic Resonance (MRI), X-Rays, Computed Tomography (CT), Ultrasound. Histopathologic Images (HI) are another kind of medical image obtained by means of microscopy of tissues from biopsies and bring to the specialists the ability to observe tissues characteristics in a cell basis. 

Cancer is a disease with a high mortality rate in developed and in developing countries. Beyond the life lost problem, the costs for health treatment are high and have an impact on government and population. According to \citet{Torre201587}, the rates of mortality among high-income countries are stabilizing or even decreasing due to programs to reduce the risk factors (e.g. smoking, overweighting, physical inactivity) and due to treatment improvements. In low and middle-income countries mortality taxes are increasing due to the increase in risk factors. Improvements in treatment include the early detection. In 140 of 184 countries, breast cancer is the most prevalent type of cancer among women \cite{Torre2017}. Imaging exams like mammography, ultrasound or CT can diagnose the presence of masses growing in breast tissue, but the confirmation of the type of tumor can only be accomplished by a biopsy. Biopsies take more time to provide a result due to the acquiring procedure (e.g. fine needle aspiration or surgical open biopsy), the tissue processing (creation of the slide with the staining process) and finally pathologist analysis. Pathologist analysis is a highly specialized and time consuming task prone to inter and intra-observer discordances \citep{BELLOCQ2011S92}. The variance in the analysis process can be caused by the staining processes also. Hematoxylin and Eosin (H\&E) is the most common and accessible stain, but they can produce different color intensities depending on the brand, the storage age, and temperature. Therefore, Computer Aided Diagnosis (CAD) can increase pathologists' throughput and improve the confidence of results by adding reproducibility and reduce observer subjectivity.

Histopathologic images (HI) have different visual characteristics comparing to traditional image processing of macro vision problems like ImageNet \cite{russakovsky2015}. Currently, deep methods like Convolution Neural Networks (CNN) and Deep Neural Networks (DNN) are gaining attention by the scientific community due to last recent results on large datasets. Training deep methods is known to need a high amount of images due to their automatic feature learning nature. Usually, such an amount of data is not available on HI. To overcome this problem two approaches can be used: data augmentation or transfer learning.

\begin{figure}
	\centering
    \includegraphics[width=3.5in]{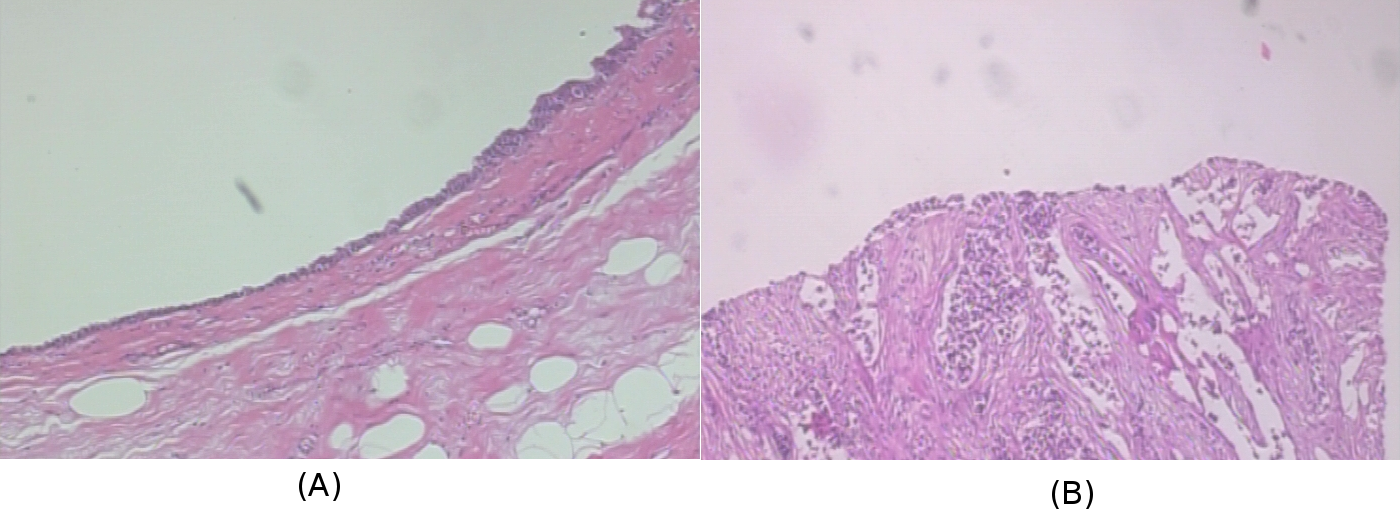}
	\caption{A - Example of a Adenosis (benign tumor) slide, B - Example of a Ductal Carcinoma (malign tumor) slide}
    \label{fig:adenoma_ductal_carcinoma_1}
\end{figure}

For data augmentation usually, images are transformed using affine transforms to avoid inserting biases on the classification process using other morphological operations. Another way to produce data augmentation is patching the images. Patching can produce the effect of selecting pieces of an image with the same structure but that pertain to images that are from different classes. An example can be seen in Figure \ref{fig:adenoma_ductal_carcinoma_1} where patches from different images from different classes seem visually similar. One important feature in cancer diagnosis is the observation of nuclei. Tumors like Ductal Carcinoma and Lobular Carcinoma presents an irregular growing on epithelial cells at these structures. A high number of nuclei or high amount of mitotic cells in a small region can indicate the presence of an irregular growth of tissue, representing a tumor. An HI can capture this feature, but besides the nuclei, it will capture other healthy tissues that can be seen in images of benign tumors. Stroma is one type of tissue that can be seen with the same characteristics in parts of malign and benign images. Selecting more relevant patches could improve the classification processes.

Deep methods learn their weights from data, so more images means a training process with more quality. ImageNet is one of these cases, despite the fact that it is a thousand class problem, the number of images for training is high (~14 million). HI datasets are usually smaller, so training from scratch CNNs like Resnet, Inception or Densenet becomes difficult as we have few data compared to the number of parameters to learn. The transfer learning method tries to use networks that were previously trained in bigger datasets to classify images in another context. This method can be applied in two ways, to extract features from images of a small dataset (e.g. HI datasets) and use these features in another shallow classifier. Another way is to use a network trained in a bigger dataset as a base for training on a small dataset. This means that filters learned from the other dataset will be the starting point to learn more from the new dataset. This process is known as fine-tuning and includes the adaptation of the classifier layer of the previous network for the new problem because usually, the number of classes is different.

This work presents a literature review related to HI and computer techniques to process this type of images. Section \ref{sec:lit} presents the Literature Review that is subdivided in Datasets, Unsupervised Shallow Methods, Supervised Shallow Methods, Deep Methods and Reviews. Section \ref{casestudy} presents a case study of a classification problem using a merge of deep methods and shallow methods to classify breast cancer HI. In Section \ref{conclusion} we present a conclusion about the Literature Review and the Case Study.

%%%%%%%%%%%%%%%%%%%%%%%%%%%%%%%%%%%%%%%%%%%%%%%%%%%%%%%%%%%%%%%%%%%%
\section{Literature Review}
\label{sec:lit}
%%%%%%%%%%%%%%%%%%%%%%%%%%%%%%%%%%%%%%%%%%%%%%%%%%%%%%%%%%%%%%%%%%%%
This section is divided into four subsections: Datasets, Unsupervised shallow methods, Supervised shallow methods, Deep methods, and Reviews. The dataset section presents some publicly available datasets can be used for the purposes of this work. Unsupervised shallow methods are dedicated to present works that used algorithms like K-Means, Gaussian Mixture Model mainly for segmentation of HI. Supervised shallow methods subsection presents works that used monolithic classifiers and ensembles to classify images based on labeled data. It also contains a subsection of feature extraction that is essential for providing data to shallow methods. Deep methods subsection presents works that use both supervised and unsupervised deep methods e.g. CNNs, DNNs, and Autoencoders. The reviews subsection brings the surveys that were found during the search. Usually, reviews are not included in systematic literature reviews, but for the purpose of this document, we judge this listing important.

Common tasks involving HI images are the segmentation and classification. Usually, segmentation is used to separate structures of the image that can help the diagnosis. One key structure that is the focus of many segmentation works is the nucleus. Pleomorphism and mitosis are two features found in histopathology images that can help to diagnose or give a prognosis to patients. Nucleus identification can be used by automated mechanisms to identify the presence of cancer or its classification. Generally, the output of the segmentation is the input image with identified areas. Usually, segmentation is an unsupervised problem (due to the time-consuming task of fine grain labeling) or uses a dataset that provides ground truth images for performance comparison.

Classification problems use images as input and provide as output a probability of an image pertaining to each class of the problem. Classes can be e.g. the type of tumor or the grade of the tumor. This process usually involves a feature extractor and a classifier. The advances in Machine Learning that lead to the emerging of Convolutional Neural Networks altered some aspects of the process. A CNN can be fed with a raw HI and provide the class probability output. Another way to proceed is by using a CNN to obtain features at the higher layers and provide this features to a shallow classifier. A common problem faced by deep methods is that unlike object identification problems the number of images on HI datasets is much smaller. So, it is difficult to train a data-dependent model like a CNN with a small number of images. To overcome this problem the fine-tuning, data augmentation and transfer learning is broadly applied to HI.

The literature review aims to answer three research questions:
\begin{itemize}
\item RQ1: Which machine learning (ML) methods are being used for HI classification and how these images are provided to the ML methods (feature extraction or the images themselves)? This RQ tries to identify which classifiers, ensembles of classifiers or deep methods are being used to classify images that are the subjects of this work. It tries to clarify how these images are processed to generate feature vectors for classification procedures.
\item RQ2: Which elements are usually classified related to these images and what are the elements of interest in these images? The objective of this question is to identify which type of tissues or structures can be identified using classification or clustering.
\item RQ3: Which image processing methods are being applied to improve classification results? This question has as purpose to identify what well-known methods can be used before feature extraction or classification of the HI 
\end{itemize}

Based on the research questions we created a search query \footnote{((histology AND image) or (histopathology AND image) or (eosin AND hematoxylin)) and (("machine learning") or ("artificial intelligence") or ("image processing"))} and used it in five research portals: IEEE Xplore, ACM Digital Library, Science Direct, Web of Science and Scopus. The period of the search was restricted between 2008 and 2018.
Table \ref{table:1} presents the number of results obtained with the search query. Each search engine used a different query based on the original. We searched based on title, abstract and keywords for all search engines, except for Science Direct. In this case, we added the full-text search also, because the number of relevant works suffered a high drop without it.
\begin{table}[htpb!]
\centering
\footnotesize
\begin{tabular}{l r r r} 
 \hline
 Search engine & Results & 1st filter & 2nd filter\\ 
 \hline
 IEEE Xplore & 96 & 68 & \\ 
 ACM Digital Library & 5 & 3 & \\
 Science Direct & 1752 & 161 & \\
 Web of Science & 409 & 54 & \\
 Scopus & 252 & 67 & \\
 \hline
 Total & 2514 & 353 & 151\\
 \hline
\end{tabular}
\caption{Number of results without exclusion criteria, after the application first exclusion criteria and after the second exclusion criteria}
\label{table:1}
\end{table}

Based on these results, the first exclusion step was based on the title and abstract. Most of the exclusion in this step was due to papers that mentioned "image processing" in the text, but the sense of the term was linked with the process of digitalizing HI images only for pathologist analysis. Another exclusion factor in the first step was the presence of the term eosin and hematoxylin or histopathology in the text only mentioning it for comparison effects, but with the actual focus of the work on CT, MRI or radiologic images.

In the last step of the first exclusion, we eliminated the duplicated articles resulting in 353 articles. The second exclusion was finer, excluding articles after reading of their content, resulting in 169 articles left.

\begin{figure}[!ht]
\centering
\includegraphics[width=3.5in]{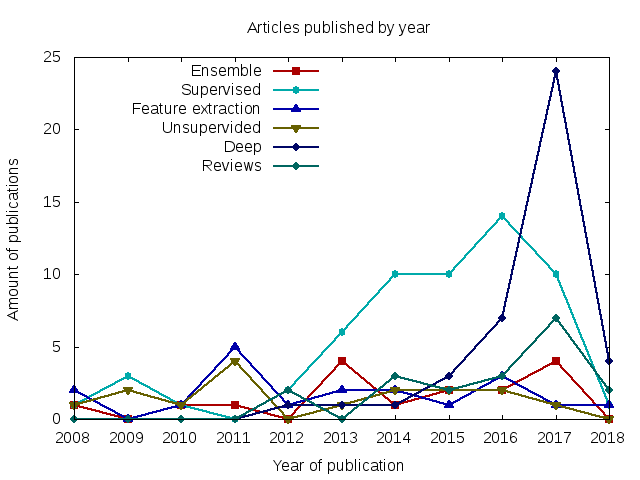}
\caption{Amount of articles per year after filters}
\label{fig:statistics}
\end{figure}

The amount of publications related to the field of this research is presented in Figure \ref{fig:statistics}. Based on Figure \ref{fig:statistics} it is possible to note that the research on the topic is increasing in last years. The search was accomplished in May of 2018, even with the search at the beginning of the year some publications on 2018 were found. It is also possible to note a great increase in the use of deep methods, while ensembles and feature extraction kept their rates. Table \ref{table:2} shows the number of publications per journal between the years 2008 and 2018 on the subject of this research. These are the top 15 journals in number of publications. The top 15 conferences are presented in Table \ref{table:3}.

\begin{table}[htpb!]
\footnotesize
\begin{tabular}{llc}
 \hline
 Journal title & Area & Pub.\\ 
 \hline
Computerized Medical Imaging and Graphics    & CHM & 5 \\
Computers in Biology and Medicine            & CM & 5 \\
Pattern Recognition                          & C & 5 \\
Computer Methods and Programs in Biomedicine & CM & 4 \\
Medical Image Analysis                       & CHM & 4 \\
Procedia Computer Science                    & C & 4 \\
Artificial Intelligence in Medicine          & CM & 3 \\
Journal of Medical Imaging                   & M & 3 \\
Neurocomputing                               & CN & 3 \\
Bioinformatics                               & BiCMa & 2 \\
Expert Systems with Applications             & CE & 2 \\
IEEE Transactions on Biomedical Engineering  & E & 2 \\
IEEE Transactions on Medical Imaging         & CEH & 2 \\
Micron                                       & Bi & 2 \\
Pattern Recognition Letters                  & C & 2 \\
\hline
\end{tabular}
\caption{Number of publications per journal between 2008 and 2018 - C=Computing, M=Medicine, H=Health sciences, N=Neuroscience, Bi=Biochemistry, Ma=Mathematics, E=Engineering}
\label{table:2}
\end{table}

\begin{table}[htpb!]
\footnotesize
\begin{tabular}{ll}
 \hline
 Conference & Pub.\\ 
 \hline
- Medical Imaging & 9 \\
- IEEE International Symposium on Biomedical Imaging & 7 \\
- SPIE & 3 \\
- International Conference on Pattern Recognition & 3 \\
- \parbox[2cm]{6cm}{\rule{0pt}{1em}International Conference of the IEEE Engineering in\\ Medicine and Biology Society\rule[-0.3em]{0pt}{1em}} & 3 \\
- Medical Image Computing and Computer-Assisted Intervention & 2 \\
- \parbox[2cm]{6cm}{\rule{0pt}{1em}International Symposium on Medical Information\\Processing and Analysis\rule[-0.3em]{0pt}{1em}} & 2 \\
- \parbox[2cm]{6cm}{\rule{0pt}{1em}Signal Processing and Communications\\Applications Conference\rule[-0.3em]{0pt}{1em}} & 1 \\
- \parbox[2cm]{6cm}{\rule{0pt}{1em}International Conference on Innovations in \\Information, Embedded and Communication Systems\rule[-0.3em]{0pt}{1em}} & 1 \\
- \parbox[2cm]{6cm}{\rule{0pt}{1em}International Conference on Information Technology\\in Medicine and Education\rule[-0.3em]{0pt}{1em}} & 1 \\
- \parbox[2cm]{6cm}{\rule{0pt}{1em}International Conference on Image Processing\\Theory, Tools and Applications\rule[-0.3em]{0pt}{1em}} & 1 \\
- International Conference on Hybrid Intelligent Systems & 1 \\
- \parbox[2cm]{6cm}{\rule{0pt}{1em}International Conference on Electronic Systems,\\ Signal Processing and Computing Technologies\rule[-0.3em]{0pt}{1em}} & 1 \\
- International Conference on Electrical Engineering and Informatics & 1 \\
- \parbox[2cm]{6cm}{\rule{0pt}{1em}International Conference on Digital Image\\ Computing: Techniques and Applications\rule[-0.3em]{0pt}{1em}} & 1 \\
\hline
\end{tabular}
\caption{Number of publications per conference between 2008 and 2018}
\label{table:3}
\end{table}

%%%%%%%%%%%%%%%%%%%%%%%%%%%%%%%%%%%%%%%%%%%%%%%%%%%%%%%%%%%%%
\subsection{Datasets}
%%%%%%%%%%%%%%%%%%%%%%%%%%%%%%%%%%%%%%%%%%%%%%%%%%%%%%%%%%%%%
Three dataset proposals were found. \citet{Kumar20171550} proposes a dataset with multiple tissue sources, where the focus is the nuclear segmentation. All images are annotated identifying nuclei of WSIs totalizing 21000 images. Besides the dataset, a framework for the evaluation of segmentation results is also proposed.

\citet{Spanhol2016} presents a dataset of 82 patients with diagnoses of eight types of breast tumors. A framework for the evaluation of the classification methods for eight types of tumor is also proposed. A baseline accuracy value is presented using six feature extractors and four classifiers. The dataset is composed of 7,909 images of four magnification factors. Images are 460 pixels height and 700 pixels wide.

A dataset of colorectal cancer is presented by \citet{Kather2016}. The proposed dataset is labeled with 8 types of structures that are commonly found on WSI. Ten large WSI were tesselated to produce 5000 images of 150x150 pixels, 625 images per structure type. The work contains a baseline for comparison purpose based on five texture descriptors and three classifiers.

\citet{Kostopoulos2017} proposed a dataset of 3,870 HI stained with Hematoxylin\&Eosin or Immunohistochemistry. This dataset contains three types of tissue origin: breast, brain, and larynx. It does not present a baseline. \cite{Kostopoulos2009}\cite{Ninos2016}\cite{Glotsos2008}

This section presented works published with the specific purpose of proposing a new dataset. Looking deeper in the other works that propose techniques to process HI is possible to find other datasets. A review presented in the Review section created by \citep{Komura2018} brings a collection of datasets that were used in recent HI works. 

%%%%%%%%%%%%%%%%%%%%%%%%%%%%%%%%%%%%%%%%%%%%%%%%%%%%%%%%%%%%%
\subsection{Unsupervised Shallow Methods}
%%%%%%%%%%%%%%%%%%%%%%%%%%%%%%%%%%%%%%%%%%%%%%%%%%%%%%%%%%%%%
\begin{table*}[!ht]
\footnotesize
\centering
\begin{tabular}{l c l l} 
 \hline
 Reference & Year & Tissue region & Method \\
 \hline
 \hline
 	\citet{Liu2008650} & 2008 & Lymphocytes & \parbox[2cm]{4cm}{\rule{0pt}{0.5em}ISODATA\\Synergic Neural Network\rule[-0.3em]{0pt}{0.5em}} \\
 \hline
	\citet{Tosun20091104} & 2009 & Colorectal & K-means \\
 \hline
	\citet{5192968} & 2009 & Prostate & \parbox[2cm]{4cm}{\rule{0pt}{0.5em}Spatial Constraint\\Fuzzy C-Means\rule[-0.3em]{0pt}{0.5em}} \\
 \hline
	\citet{5600019} & 2010 & Cervix & GMM \\
 \hline
	\citet{5415659} & 2010 & Breast & K-means \\
 \hline
 	\citet{Roullier2011603} & 2011 & Breast & K-means \\
 \hline
	\citet{6061451} & 2011 & Uterus & K-means \\
  \hline
	\citet{5872824} & 2011 & Prostate & K-means \\
 \hline
	\citet{6061453} & 2011 & Uterus & K-means \\
 \hline
 	\citet{Onder201333} & 2013 & Colorectal &  \parbox[2cm]{4cm}{\rule{0pt}{0.5em}Quasi-supervised\\ Nearest Neighbors\rule[-0.3em]{0pt}{0.5em}}\\
 \hline
 	\citet{Nativ2014228} & 2014 & Liver & K-means \\
 \hline
 	\citet{Yang2014996} & 2014 & Prostate & Mean shift + similarity \\
 \hline
	\citet{ISI:000344338900003} & 2014 & Brain & K-means \\
 \hline
 	\citet{Sirinukunwattana201516} & 2015 & Breast & Dictionary/thresholding \\
 \hline
 	\citet{Huang2015} & 2015 & Breast & Mean-shift clustering \\
 \hline
 	\citet{7976445} & 2016 & Lymphocytes & K-means \\
  \hline
	\citet{Mazo20161} & 2016 & Cardiac & K-means \\
  \hline
	\citet{ISI:000382313300034} & 2016 & Lung & K-means \\
  \hline
  	\citet{Shi2017} & 2017 & Liver & K-means \\
  \hline
	\citet{Shi201799} & 2017 & Lymphocytes & K-means \\
  \hline
  	
\end{tabular}
\caption{Summary of publications focused on unsupervised methods}
\label{table:unsupervised_table}
\end{table*}

K-means is a unsupervised machine learning method and is the core or base for twelve works found in this survey. Table \ref{table:unsupervised_table} shows the works under the unsupervised category. \citet{5415659} proposed a method based on Expectation-Maximization of Geodesic Active Contour to detect lymphocyte nuclei. Four structures can be identified, lymphocyte nuclei, stroma, cancer nuclei, and background. The process initiates with a K-means segmentation, this segmentation is improved with an Expectation-Maximization model. The contours are identified based on magnetic interaction theory. After contour have been defined, an algorithm searches for the concavity of contours, meaning there are nuclei overlapping. They used a breast cancer dataset. K-means is applied in \citet{6061453} to help on the classification of HI, although the focus is not on the K-means, but on Gabor filters, this clustering method is essential in the segmentation process. \citet{5872824} used K-Means and PCA to divide the image into four types of structures: glandular lumen, stroma, epithelial-cell cytoplasm and cell nuclei. Subsequently, they performed morphological operations of closing and filling. \citet{6061451} used a merging of local region-scalable fitting and K-means to produce the segmentation of Cervix images. \citet{ISI:000382313300034} presents an approach for segmentation that starts with a K-means segmentation. The result of this prior segmentation is then improved and simplified using a sequence of thresholds that tries to keep the form of objects. A Random Forest algorithm used the visual information based on the coloring of neighbors to identify the nuclei. They used one more classification step by training a new Random Forest algorithm now with good examples of segmented nuclei in the previous process. A feature set of elliptic fit, shape index and width x height ratio are the source for the second Random Forest. The key point of the article is not actually the segmentation, but the nuclei detection. \citet{Shi201799} proposes a segmentation method considering local correlation feature of each pixel. They used two K-means clustering, the first K-means applied generated a poor segmented cytoplasm, a second K-means does not consider the nuclei identified by the first clustering. The final step is a watershed to finalize the segmentation. \citet{ISI:000344338900003} uses K-means for segmentation of HI. Skeletonization and Shock Graphs identified nuclei in the previously segmented image. If the identification of nuclei using a Shock Graph provides a confidence value minor then 0.5, it performs another classification phase using a MLP. This hybrid approach achieves 92.5\% of accuracy, the same as the best comparable result.

The work proposed by \citet{Liu2008650} uses the ISODATA algorithm to cluster images of cells to create prototypes. Synergetic Neural Network classified these prototypes. They used morphological features (size, shape, lucency and texture roughness). \citet{Tosun20091104} proposes a segmentation based on K-means. K-means clusters all pixels into three categories (purple, pink, white) and a thresholding procedure separate each category into three subcategories. The object-level segmentation based on clustering achieved 94.89\% of accuracy against 86.78\% for pixel-level segmentation. \citet{5192968} studies two strategies for initialization of clustering methods: geodesic active contours and multiphase vector level sets. The last one proved to be more efficient when using Spatial Constraint Fuzzy C-Means, with accuracy values of 68.1\% and 67.9\% respectively. K-Means achieved 60.6\% in this case. \citet{5600019} presents a segmentation based on Gaussian Mixture Models. It uses the stain color features (hematoxylin with blue color and eosin in pink and red) to apply two segmentation steps on R channel and other channels subsequently. It does not present ground truth comparison, only visual results comparing to K-means.

A multi-scale segmentation with K-means is the subject in \citet{Roullier2011603}. This work uses the same idea of the pathologist to analyze a WSI. The segmentation starts at a lower magnification factor and finishes at a higher magnification where it is easier to identify mitotic cells. The result of the clustering algorithm aims to identify regions of interest at each magnification. A Quasi-supervised approach, based on nearest neighbors, can be used to cluster one unlabeled dataset based on itself and another labeled dataset \cite{Onder201333}. A comparison between Quasi-supervised approach and SVM, using GLCM features, show that SVM provides better performance but requires labeled data. \citet{Nativ2014228} presents a K-means clustering based on morphological features of lipid droplets previously segmented using Active Contours Model (ACM). A Decision Tree was used to verify the rules that lead to the classes obtained by the clustering. The correlation with the pathologist evaluations achieved 97\%.

\citet{Yang2014996} proposes a system for content recovery. It is three steps method that uses histograms features. The first two steps use dissimilarity measures of histograms to find candidates. The last step uses mean-shift clustering. The AUC of the proposed method is 0.87, which is better than 0.84 of the LBP based method. A mitotic cell detection system using a dictionary of cells is presented by \citet{Sirinukunwattana201516}. A shrinkage/thresholding method groups intensity features represented by a sparse coding to create a dictionary. It achieved 0.848\% of accuracy on MITOS dataset. The separation of stains can improve the results from analysis of HI. To address this problem, \citet{Huang2015} proposes a semi-supervised method based on Independent Component Analysis (ICA) called Exclusive Component Analysis (XGA). The difference between this method and the others from literature is the need for a small interference from the user, which must provide a set of references from nuclei and from the cytoplasm.

A two-step K-means is used by \citet{7976445} in order to segment follicular lymphoma HI. The first step segments nuclei and another type of tissues in a two-class cluster. The next step segments areas from the previous step that are not nuclei into three classes (nuclei, cytoplasm and extracellular spaces). The final step is a watershed algorithm to get better contours of nuclei. The difference between the manual segmentation and automated was around 1\%. \citet{Mazo20161} uses K-means to segment cardiac images in three classes: connective tissues, light areas, and epithelial tissue. A flooding algorithm processes light areas in order to merge its result with epithelial regions and improve the final result. Finally, plurality rule classifies cells into flat, cubic and cylindrical. This method achieved a sensitivity of 85\%. This work is extended in \citet{Mazo20171}. K-means is used to cluster pixels represented by the L*a*b color space over color and neighborhood statistics in \citet{Shi2017}. A thresholding step finally improves contours detection of fat droplets. Specialists evaluate the morphological information related to the droplets to analyze the results.

%%%%%%%%%%%%%%%%%%%%%%%%%%%%%%%%%%%%%%%%%%%%%%%%%%%%%%%%%%%%%

\subsection{Supervised Shallow Methods}
\label{sec:supervised}

%%%%%%%%%%%%%%%%%%%%%%%%%%%%%%%%%%%%%

\subsubsection{Classification}

Classification can identify types of tumors, types of tissues, nucleus features (e.g. mitosis phases) or specific characteristics from some organs (e.g fat inside liver or the size of epithelial tissue on the cervix). Table \ref{table:classification_table} summarizes works that deal with the classification task. In this section, we can see a more diversified use of classifiers, with 16 works using SVM and seven works using Random Forests from 33 works.

\begin{table*}[!ht]
\centering
\footnotesize
\begin{tabular}{l c l l} 
 \hline
 Reference & Year & Tissue region & Classifier \\
 \hline
 \hline
 	\citet{Marugame2009173} & 2009 & Breast & Bayes classifier \\
 \hline
 	\citet{Mete2009284} & 2009 & Skin & Naive Bayes, SVM, Tree \\
 \hline
 	\citet{5505922} & 2010 & Lymphoma & RBF network, Naive Bayes and KNN \\
 \hline
 	\citet{Osborne:2011:MCM:1982185.1982210} & 2011 & Skin & SVM \\
 \hline
 	\citet{Sidiropoulos2012376} & 2012 & Brain & Probabilistic Neural Network \\
 \hline
 	\citet{6353218} & 2013 & \parbox[2cm]{2.5cm}{\rule{0pt}{0.5em}Multiple organs \\ and tissues \rule[-0.3em]{0pt}{0.5em}} & KNN + Late fusion \\
 \hline
 	\citet{6450064} & 2013 & Breast & Random Forests\\
 \hline
 	\citet{De2013475} & 2013 & Uterus & LDA\\
 \hline
 	\citet{Homeyer2013313} & 2013 & Liver & Random Forests, KNN and Naive Bayes \\
 \hline
 	\citet{Cosatto2013} & 2013 & Gastric & MLP and Multi instance learning \\
 \hline
 	\citet{Vanderbeck2014785} & 2014 & Liver & SVM \\
 \hline
 	\citet{6868127} & 2014 & Gastric & \parbox[2cm]{4cm}{\rule{0pt}{1em}MIL (SIL-SVM, MI-SVM,\\ mi-SVM and mi-Graph) \rule[-0.3em]{0pt}{1em}}\\
 \hline
 \parbox[2cm]{3.5cm}{\rule{0pt}{0.5em}\citet{6943991} and \citet{7367154}\rule[-0.3em]{0pt}{0.5em}} & 2014 & Liver & SVM \\
 \hline
 	\citet{Irshad2014390} & 2014 & Breast & MLP, SVM, Tree \\
 \hline
  	\citet{Xu2014591} & 2014 & Colorectal & cMIL (constrain-based MIL) \\
 \hline
	\citet{7371235} & 2015 & Colorectal & KNN \\
 \hline
 	\citet{Korkmaz20154026} & 2015 & Breast & LSSVM \\
 \hline
 	\citet{Tashk20156165} & 2015 & Breast & SVM and RF \\
 \hline
 	\citet{Kandemir201544} & 2015 & Gastric & MIL (benchmarking of multiple methods) \\
 \hline
 	\citet{ISI:000367870000001} & 2015 & Lymphoma & SVM (ensemble) \\
 \hline
 	\citet{Gertych2015197} & 2015 & Prostate & SVM and Random Forests \\
 \hline
 	\citet{ISI:000399823502195} & 2016 & Breast & Random Kitchen Sink \\
 \hline
 	\citet{ISI:000383210600025} & 2016 & Breast & Random Forests \\
 \hline
	\citet{Wright2016125} & 2016 & Colorectal & Random Forests \\
 \hline
    \citet{AngelArulJothi2016652} & 2016 & Thyroid & VPRS + CMR \\
 \hline
 	\citet{Barker201660} & 2016 & Brain & Elastic Net \\
 \hline
 	\citet{Pang2017} & 2017 & Liver & RF, SVM, ELM (Concave-Convex Variation) \\
 \hline
 	\citet{Mazo20171} & 2017 & Cardiac & SVM Cascading \\
 \hline
 	\citet{ISI:000404037600002} & 2017 & Breast & Random Forests \\
 \hline
 	\citet{Peikari20171078} & 2017 & Breast & Cascade SVM \\
 \hline
 	\citet{BenTaieb2017194} & 2017 & Ovary & LSVM \\
 \hline
 	\citet{Zhang201744} & 2017 & Lung & SVM + Sparse coding \\
 \hline

 \hline
 	\end{tabular}
\caption{Summary of publications focused on classification}
\label{table:classification_table}
\end{table*}

\citet{Marugame2009173} proposes a simple classifier based on Bayes theory to classify ductal carcinomas into three categories. A preprocessing step based on morphological operations and Gabor Wavelets segments images. Morphological features represent the images for classification. Specialists consulted in the article claimed that the simple classifier provides, together with the morphological features, a better way to understand the results.

A comparison of the impact of color representation is presented in \citet{Mete2009284}. This work tests eleven different color spaces for representing HI and from each one the feature extraction step extracts features called positive and negative colors. The combination of a Spherical Coordinates Transform and Decision Tree showed to be the best in this case. \citet{5505922} also compares four color spaces, i.e. RGB, L*a*b, grayscale and RGB with Eosin and Hematoxylin representation. They used eleven features (e.g. Zernike, Chebychev, Histograms, GLCM and Edge statistics) to represent the images. Weighted KNN obtained the best results (claimed at 99\%) followed by the RBF network and Naive Bayes with 99\% and 90\% respectively. A colorspace called Eosin representation obtained the best results.

\citet{Osborne:2011:MCM:1982185.1982210} uses 4 morphological features extracted after segmentation to create a feature vector for a SVM classifier. The highlight is the selection of most relevant features. The nuclei irregularities (rate between the major and minor radius and rate between area and perimeter) can provide better results achieving 90\% of accuracy.

\citet{Sidiropoulos2012376} proposes a classification algorithm based on Probabilistic Neural Network implemented using GPU. The advantage is the reduced processing time that allows an exhaustive feature combination search. For demonstration purposes, a comparison of CPU and GPU based algorithms showed that the GPU version takes 278 times less computation time than the CPU version for a feature vector with 20 attributes.

KNN with late fusion is used by \citet{6353218} to classify tissues from two datasets of HI of different organs. The focus is on the selection of features using an objective function implemented by KNN. A set of seven feature extractor provides the information to be combined. 

A multi-field-of-view (FOV) classification scheme is proposed in \citet{6450064}. It uses a multiple patch size procedure for WSI. The work firstly analyses what features (morphological, textural or graph based) are more relevant at each patch size. After that, it uses a Random Forests based classification scheme using the aggregation of multiple FOV patches. It does not present a baseline or accuracy comparison, only the AUC result, showing better values for nuclei architecture features to recognize low versus high-grade ductal carcinoma.

\citet{De2013475} proposes grading of uterine cervical cancer using an LDA classifier. It uses a 62-wide feature vector based on textural (GLCM), graph (Delaunay-triangulation) and morphological metrics. A specialist manually segmented the images for classification to identify the tumor and divided the tumor into ten segments for feature extraction. A voting strategy combined results from the segments. The whole epithelium best result is 54\% against 62\% for the vertically segmented epithelium. An extension of this work \cite{7307080} presented an automatic orientation detection for the epithelium with more features and used an SVM classifier.

WSI is the core of the work proposed in \cite{Homeyer2013313}. It compares KNN, Naive Bayes and Random Forests for classification of slides based on a patching procedure. The features used were LBP, intensities histograms and a combination of them. Random Forests with a group of all features obtained the best result (94.7\%).

A stain separation is performed in \citet{Cosatto2013} using an SVR regressor to identify a high occurrence of hematoxylin in low-level magnification, meaning that it is a region of interest (ROI). A ridge filter binarizes the ROIs of the images at higher magnifications (100x and 200x) allowing to measure nuclei and generate morphometric features. This work uses a Multi-Instance Learning (MIL) approach because ROI is not labeled, only the whole slides are, so all ROIs from a positive slide to receive positive labels. MIL uses MLP for classification, but it requires a modified loss function to represent the one-positive rule for a slide, which means that whether in the prediction only one ROI appears as positive, all slide is positive. The comparison between the MIL approach and SVM classification. The SVM required ROI labeling. The AUC of MIL was 95.6\% against 94.9\% of SVM with the advantage of reducing labeling efforts.

In \citet{Vanderbeck2014785} an SVM classifies white regions of liver HI among 7 classes. It uses morphological, textural and pixel neighboring statistics features. The best accuracy was of 93.5\% for all features combined in a 413-wide feature vector. The work also compares the results based on labeled images by different pathologists.

The MIL method is applied in \cite{6868127}. It compares three MIL SVMs, i.e. SIL-SVM, MI-SVM and mi-SVM with mi-Graph. The last one outperforms with 87\% against 69\% on the best one (mi-SVM). Their work is based on patching and uses cell-level morphometric features and patch-level textural, color and morphometric features. All images are previously segmented using the watershed algorithm. Another work from the same research group performs a benchmark of MIL methods, finding out that MILBoost gives better accuracy results on instance-level approach (66.7\%) and mi-Graph performs better in bag-level prediction (72.5\%) \cite{Kandemir201544}.

\cite{6943991} proposes a feature selection method based on morphometric, textural and structural features. A greedy algorithm (FSelector and In-house recursive) selected features in a pool of 200 features. The selection's objective function is an SVM classifier. They performed the classification using Tree and SVM classifiers, where the first one with In-house recursive algorithm achieved 95.7\% of accuracy with 20 features. Improvements in the work were published in \cite{7367154} with the addition of more descriptors.

The work presented in \cite{6830728} classifies follicular lymphomas. It uses a preprocessing step to segment images based on intensity thresholds and an expectation maximization algorithm. Linear Discriminant Analysis classifies the resulting cells from the segmentation. It presents a detection rate of 82.6\%, but the work does not show a baseline for comparison.

A multimodal approach with multispectral images based on MITOS 2012 contest dataset focused on selecting the best spectral bands for classification of mitotic cells \cite{ }. The method generates a ranking of the best bands using intensity, morphological and textural features from each channel. SVM, Tree, and MLP are the classifiers used. SVM obtained the best result (63.7\%) using only the 8 best bands, as a comparison, the best RGB value is 52\% with SVM also. The best state-of-art for the time of the publication was 59\%.

A MIL method called MCIL used a patching procedure to create instances for MIL classification \cite{Xu2014591}. A Boosting ensemble method with Gaussian classifier classified the patches clustered using K-means. The work performs comparisons with regular image-level classification methods and MIL methods. The fully supervised method presented 76.6\% (using patch labeling) and the proposed method achieved 69.9\%. MCIL achieved 71.7\% and 60.1\% in another dataset (not patch labeled) with constrained and unconstrained MCIL respectively against 25.3\% of MIL-Boosting.

A colorectal CAD system is presented in \cite{7371235}. The method starts with an Otsu thresholding of red channel to separate nuclei, background, and stroma. The features for the images are the ratio between nuclei and stroma inside small patches of images (nucleus features). A p-type Fourier descriptor, extracted after thresholding of the cells, represents the glands contours. Another feature is the condition of glands between atypical and regular obtained by a KNN classifier using the glands contours. An SVM classifier produces the final classification results which obtained 78.3\% of accuracy against 67\% of a GLCM based method.

\citet{Korkmaz20154026} proposes a classification framework based on minimum redundancy maximum relevance feature selection and Least Square SVM. It is claimed an accuracy of approximately 100\% with only four false negatives for benign tumors in a three-class problem. They do not perform more comparisons.

A complete framework since segmentation, feature extraction, feature selection, and classification is presented by \citet{Tashk20156165}. It uses Maximum Likelihood Estimation to obtain the mitotic pixels in L*a*b color space. Mitotic candidates have their features extracted using Completed Local Binary Pattern, morphometric features, statistical features and stiffness matrix. A cascading classification is performed firstly with SVM and secondly with RF. A comparison shows that this method produces the accuracy of 96.5\% against 82.4\% of the best previous result on MITOS 2012 dataset.

\citet{ISI:000367870000001} uses SVM in two steps of the classification. Firstly, it is used to perform the segmentation of the WSI images. The segmentation identifies regions that can be considered regions of interest. The second step is a classification using an ensemble of SVMs. The accuracy of multiple "weak" classifiers trained with fewer features and different parameters is used for the final result, combining them with a weighted sum function. The best result was an accuracy of 88.6\%.

\citet{Gertych2015197} presents a system for prostate cancer classification. It consists of two classifiers, SVM and Random Forests. The first one separates the stroma and epithelium and the last one for benign/normal and cancer tissue identification. Feature descriptors were histograms of H\&E stains, rotation invariant local variance, and LBP. The best accuracy result was 68.4\% for cancer detection.

A Random Kitchen Sink (RKS) classifier is used by \citet{ISI:000399823502195} to identify mitotic nuclei on breast cancer HI. Nuclei are identified using thresholding on the red channel. Local Active Contour Model selects and models nuclei. RKS uses the intensity, shape and textural (GLCM) features from nuclei and for classification. On MITOS 2014 dataset it achieves F1-score of 96.07 for RKS and 83.4 for Random Forests.

\citet{ISI:000383210600025} is based on the work presented in \cite{ISI:000337288300001}. Simple Linear Iterative Clustering (SLIC) extract patches by tesselation without the square shape constraint. A set of 16,128 features derived from multiple histograms and LBP (multiple radii) using L*a*b, grayscale and RGB color spaces represents each patch. This number of features is suitable for a Random Forests classifier which obtained 79.51\% of F1-score for tessellated patches, comparing to 77.57\% of squared patches and 71.80\% of the base work. Simple Linear Iterative Clustering (SLIC)

SLIC is also applied by \citet{Wright2016125} in a pipeline for colorectal cancer to initially segment images. These images have their histogram features extracted using HSV color space, and also statistics from Hematoxylin and Eosin channels were extracted. It also uses GLCM as features. A comparison showed that the proposed work achieved 79\% against 75\% from their previous work for Random Forests classification.

A CAD system proposed by \citet{AngelArulJothi2016652} used images initially converted to grayscale giving priority to the red channel. Particle Swarm Optimization (PSO) guided Otsu segments the greyscale images and the noise in segmentation is reduced using area constraints based on the nuclei size. A Closest Matching Rule (CMR) classifier classifies the reduced feature set of GLCM features preprocessed by Variable Precision Rough Sets (VPRS). It achieves results of 100\% against 99.5\% for Naive Bayes.

\citet{7455837} studies the classification of mitosis on breast cancer on a dataset labeled with four classes correspondents to the four major phases of mitosis. Classes are imbalanced, posing a challenge for the classification. The work proposes a data augmentation method for equalizing data based on PCA and its eigenvectors, comparing them to the SMOTE method.

\citet{Barker201660} used a patching procedure based on a grid over the WSI to grade brain cancer. Each patch has more general features extracted and is clustered using K-means. The final classification is performed over the features of the nuclei identified in the clustering step using an Elastic Net.

A CAD system for lung cancer detection is proposed by \citet{Pang2017}. It uses textural features (LBP, GLCM, Tamura, and SIFT), global features and morphological features. Sparse Contribution analysis selects all these features by importance. The tests used three classifiers, i.e. SVM, RF and Extreme Learning Machine (ELM) on no redundant features. Another contribution of the work is a Concave-Convex Variation. It consists of measuring the concavity of all nuclei in an image and use this value as a modifier to the probabilities of the classifier. This method presented the accuracy of 98.74\%, better than 97.68\% of the best algorithm compared. The proposed method has the advantage of being used by all classifiers, no only with RF as the compared method.

\citet{Mazo20171} proposes the classification of cardiac tissues into five categories using a block (or patch) approach. The work demonstrates the size that best represents tissues. The classification using cascading of SVM uses LBP and LBPri (rotation invariant) features from patches. The cascading firstly uses a linear SVM to separate tissues in four classes and a polynomial SVM to classify one of these classes in two sub-classes.

\citet{ISI:000404037600002} presents a system for classification of WSI. The segmentation step uses Otsu, morphological operations and histological constraints. The tests included Random Forests, SVM, KNN and Logistic Regression using 104 types of features (e.g. textural, morphometric, statistical) from random patches of segmented images. RF obtained the best accuracy result (93\%). RF also allowed discovering most important features, that includes LBP and its variations and number of nuclei.

A nuclei segmentation pipeline proposed by \citet{Peikari20171078} starts with an image segmentation based on color space changing from RGB to L*a*b followed by a multi-level thresholding and watershed algorithm. The nuclei classification uses a cascade SVM approach with LBP, histograms, GLCM and spatial features. The cascade phase initially separates lymphocytes from epithelial tissue and then classify epithelial in benign and malign. An interesting comparison is provided based on two pathologists' evaluation, the agreement between pathologists was 89\% and between the automated system is 74\% and 75\%.

The classification of ovarian cancer is the subject in \cite{BenTaieb2017194}. The proposed method localizes regions of cancer in WSI using a multi-scale mechanism considering that each tumor type has a characteristic better detected at a determined scale. The method automatically selects an ROI based on the set of data from multiple scales. The latent variable of the LSVM used for classification is the presence of a patch at a particular scale on the classification of that region. The features are bags-of-words of RGB descriptors created by a K-means clustering. The comparison of bag-of-words and LSVM approach with a deep method and handcrafted features showed that it outperforms even CNNs by 26\% and Deconvolution Networks by 15\%, achieving an accuracy of 76.2\%.

In \cite{Zhang201744} a multi-scale classification is proposed. It uses sparse coding implemented by means of Fisher discriminant analysis. An SVM classifies bags-of-words features generated by the sparse coding of SIFT features. The method outperformed the best state-of-art comparable approach with an accuracy of 81.6\% against 79.5\%.

%%%%%%%%%%%%%%%%%%%%%%%%%%%%%%%%%%%%%

\subsubsection{Ensemble}

Ensembles are a set of, usually, weak classifiers that have their results combined to provide predictions for a problem \cite{Opitz1999}. Although ensembles perform classification, they have an entire field of research in computing, so decided to make a subsection for this topic.

\citet{Daskalakis2008196} uses a pre-processing step of segmentation to enhance nuclei and extract morphological and textural features from them. A multi-classifier created using KNN, Linear Least Squares Minimum Distance, Statistical Quadratic Bayesian, SVM, and PNN uses these features for classification. The aggregation functions of classifiers predictions results are the majority vote, minimum, maximum, average, and product. A combination of features was tested and the best one for each classifier was selected. The best overall accuracy without ensemble was 89.6\% obtained by PNN classifier. For the ensemble method, the best result was 95.7\% with the majority vote.

The method proposed by \citet{Kong20091080} classifies Neuroblastomas using textural and morphological features. It considers that pathologists use morphological features for their analysis and, for computers, textural features can be easily detected. The work uses four GLCM features and a feature selection method implemented by Sequential Floating Forward Selection (SFFS). The ensemble is composed by KNN, LDA, Bayesian and SVM classifiers and uses combination by weighted voting. The scale of tested images also changes each time the classifiers cannot provide trustful results leading to a multi-resolution method.

A comparison between different ensembles approaches is presented in \cite{DiFranco2011629}. The objective of the work is to classify patches of WSI. A set of 114 features were selected and ranked using the underneath idea of Random Forests. Based on the selected and ranked features, multiple SVM (RBF and Linear) and RF classifiers were built. The aggregation function was majority voting. The results achieved were 95.5\%, 95.1\% and 94.8\% for RBF SVM, RF and Linear SVM respectively. The best prior result was 93.5\%.

\citet{5693834} proposes an ensemble of Principal Component Classifiers (PCC) called Collateral Representative Subspace Projection Modeling (C-RSPM). This ensemble classified patches, 25 of each image, represented by 50 features from the 505 extracted selected by chi-square ranking. The aggregation function is a majority vote. The performance achieved for a liver dataset was 96.41\% against 95.09\% of a KNN (k=3) and 99.4\% against 92.08\% of Adaboost on lymphoma classification.

A CAD system composed of a staining separation module, densitometric and texture feature extraction and an AdaBoost classifier is proposed in \cite{Wang20131383}. It presents a comparison with other classifiers and raw H\&E. The system obtained accuracy of 94.37\% against 86.44\% of best result of KNN classifier with the raw H\&E image.

The system described in \cite{Gorelick20131804} uses a segmentation pre-processing step to identify superpixels. An Adaboost classifies the segmented images represented by morphometric and geometric features. The system achieved an accuracy of 85\%.

A framework for cytological analysis is presented by \citet{Filipczuk20131748}. Morphometric features represent nuclei obtained after segmentation with four different algorithms, i.e. K-means, Fuzzy c-means, Competitive learning and GMM for comparison purposes. The proposed method uses a combination of Random Subspaces with one-layer perceptron to create an ensemble. The comparison showed that the proposal reached 96.0\% in comparison with 90\% of Boosting.

\citet{ISI:000313984400007} proposes a nuclei detection method based on two Adaboost stages. The first one classifies pixel-wise extracted features from stain separated images. The second Adaboost refines the result of the first with line based features. An Optimal Active Contour refines the results from the second ensemble achieving an accuracy of 95.02\% 

\citet{6974021} proposes an ensemble of "segmentors" composed of a series of Otsu thresholding algorithms with certain constraints and morphological operations. A set of 4 segmentors perform the complete segmentation, but each image can have characteristics that would require different parameters for the segmentors set. The final result of segmentation is one among 18 segmentors sets that have most parameters shared with the set of segmentors that presented less difference in the segmentation. Results showed that their approach reached 84.3\% at best against 76.4\% of other compared method.

\citet{ISI:000354372500019} used a ensemble of SVM classifiers. Each model is trained with a variation of the image prepreprocessed by Gaussian filters and color spaces. The result is an average of classifiers'  results. The AUC best value was 0.978 with no comparison presented.

\citet{ISI:000380467800126} proposes a features selection method to improve SVM-RFE. The method uses the entropy of a feature in relation to a class as a redundancy criterion and constraints in the relation between the class and feature entropy and the inter-feature entropy. An ensemble of SVM classifiers specialized in one subtype of tissue derived from prior segmentation had their result aggregated using sum rule. The performance (94.08\%) was 0.2\% better than the best SVN-RFE method using 37 features instead of 46 from the compared method.

A comparison of multiple classifiers and features is presented by \citet{7849887}. A set of monolithic classifiers is compared with Adaboost implemented with SVM, Tree, and RF. Adaboost obtained 97.8\% of accuracy.

\citet{FernandezCarrobles201699} presents a classification framework for WSI with a Bagging of decision trees with GLCM features. The best results were 99.52\% for AUC and 98.13\% for TP.

A multiview (multi-resolution) is presented by \citet{Kwak201791}. Features extracted in multiple resolutions generated the multiview approach. Four views generated 670 features and more the whole image generated 175 features of morphological and intensity values. The boosting algorithm combined the features from multiple views for classification with Linear SVM classifiers. The multiview approach achieved 98\% of AUC compared to 96\% of the concatenation of views.

\citet{Kruk2017357} uses morphometric, textural and statistical (histogram) features to describe nuclei for classification. Genetic algorithm and Fischer methods selected the most important features. The ensemble is composed of SVM and RF classifiers and trained with a subset of features from the feature selection. The best result for the method is an accuracy of 96.7\%, better than the previous state-of-art (93.1\%) and 91.1\% of a single SVM.

The Adaboost ensemble is used in \cite{ISI:000399332700026} to grade skin cancer. The ensemble classifies images described by features create with graph theory to represent the nuclei distribution. The results of accuracy achieved 72\% comparing to ground truth. 

\begin{table*}[!ht]
\centering
\footnotesize
\begin{tabular}{l l l l} 
 \hline
 Reference & Tissue region & Classifier & Agg. function\\
 \hline
 \citet{Daskalakis2008196} & Thyroid & \parbox[2cm]{4cm}{\rule{0pt}{1em}KNN\\LLSMD\\SQ-Bayesian\\SVM\\ PNN\rule[-0.3em]{0pt}{1em}} & \parbox[2cm]{3cm}{\rule{0pt}{1em}Voting\\Min\\Max\\Sum\\Product \rule[-0.3em]{0pt}{1em}}\\ 
 \hline
  \citet{Kong20091080} & Neuroblastoma & \parbox[2cm]{4cm}{\rule{0pt}{1em}KNN\\LDA\\Bayesian\\SVM\rule[-0.3em]{0pt}{1em}} & Weighted voting \\
 \hline
 \citet{5693834} & \parbox[2cm]{2cm}{\rule{0pt}{1em}Liver\\Lymphocytes\rule[-0.3em]{0pt}{1em}} & PCC & Weighted voting \\
 \hline
 \citet{DiFranco2011629} & Prostate & SVM and RF & Voting \\
 \hline
\citet{Wang20131383} &Lung & Adaboost & \\
 \hline
 \citet{Gorelick20131804} & Prostate & Adaboost & \\
 \hline
 \citet{Filipczuk20131748} & Breast & SVM + Random Subspaces & Perceptron \\
 \hline
 \citet{ISI:000313984400007} & Breast & Adaboost & \\
 \hline
 \citet{6974021} & Uterus & Set of segmentors (Otsu) & Similarity \\
 \hline
 \citet{ISI:000354372500019} & Prostate & SVM & Average \\
 \hline 
 \citet{ISI:000380467800126} & Prostate & SVM-RFE + SVM & Sum \\
 \hline
 \citet{7849887} & Prostate & Adaboost & \\
 \hline
 \citet{FernandezCarrobles201699} & Breast & Bagging and Tree & Sum and variance \\
 \hline
 \citet{Kwak201791} & Prostate & Boosting + SVM & \\
 \hline
 \citet{Kruk2017357} & Kidney & SVM + RF & Random Forests \\ 
 \hline
 \citet{ISI:000399332700026} & Skin & Adaboost \\ 
 \hline
 \end{tabular}
\caption{Summary of publications focused on ensembles}
\label{table:ensemble_table}
\end{table*}

%%%%%%%%%%%%%%%%%%%%%%%%%%%%%%%%%%%%%

\subsubsection{Segmentation}

Segmentation is one of the tasks employed in HI. It can make some structures like nuclei or glands more evident for a subsequent manual or automated classification. A list of the works focused on segmentation can be seen in Table \ref{table:segmentation_table}. SVM is the most used method of machine learning for segmentation, 7 out of 12 works use it. 

\citet{Yu2008635} uses a Spacial Hidden Markov Model (SHMM) to segment gastric HI annotated by specialists. The feature vector is obtained by Gabor filters, generating the Gabor energy values.

Structured SVM segments nuclei of images represented by a feature vector of histograms of the grayscale image submitted to morphological operations \cite{ISI:000371029500043}. This approach got 87\% against 70.2\% from previous works. The novelty in the approach is the use of intensities maps to correct segment overlapping nuclei.

\citet{Janssens20131206} presents a segmentation procedure for identifying muscular cells. It uses a first step segmentation based on thresholding. An SVM classifies the separated tissues in 3 classes, i.e. connective tissue, a clump of cells and cells. It identifies immediately connective tissues and cells. The classifier receives the patched clumps again recursively until only connective and cell tissues appear. The F1 measure is 0.62 against 0.46 of the best work compared.

\citet{Saraswat201444} proposes a segmentation procedure with NSGA-II and a threshold classifier. NSGA-II generates the threshold for features values with ground-truth images. The comparison between learned thresholds and feature values generates the segmentation.

Breast cancer prognosis is the subject in \cite{6999158}. The main objective of the article is to segment HI to facilitate classification. An SVM performs the pixel-wise classification for segmentation. This procedure separates nuclei from the stroma. A second step based on a watershed algorithm identifies nuclei. The system achieves 72\% at best using all features (pixel-level, object-level, and semantic-level).

A segmentation method to help pathologists in the WSI images analysis based on KNN is proposed by \citet{Salman2014295}. WSI can reach 30.000x30.000 pixels, so the method helps to select ROI for the pathologists. The segmentation with KNN uses histograms of 64x64 pixels patches from H\&E channels obtained by previous color deconvolution. The best result is 87.1\% for using histograms of both H\&E.

In \cite{Chen2015} a method based on pixel-wise SVM identifies stroma and tumor nests. Nuclei segmentation is carried out by a watershed algorithm. The segmentation is used to extract pixel, object and semantic-level features in a total of 400. They were reduced to 14 features by a univariate survival analysis. An analysis correlates the features to prognosis time.

\citet{ISI:000380546000311} proposes an algorithm to improve segmentation based on a watershed pre-segmented image. The cells detected by the watershed algorithm have their shape features extracted using shape measures (histogram of contours' distances from the centroids). A hyperplane compares features to identify real cells than false positives detected by watershed similarly to SVM.

A Normal Density-based Quadratic Classifier segments colorectal images in \cite{Geessink2015}. The segmentation uses L*a*b color space with a threshold to eliminate the background and HSV color space to classify the rest of the pixels. After classification, errors are corrected based on histological constraints. The algorithm showed an error of 0.6 and pathologists 4.4\% for tumor quantification.

\citet{Wang20161} proposes the use of image processing methods (wavelet decomposition, region growing, double strategy splitting model and curvature scale space) to highlight nuclei that will be classified. Nuclei have textural and shape features extracted totaling 142 features. A features selection method based on genetic algorithms and SVM is performed to obtain the best feature set. Best result was 91.5\% compared to 60.1\% of literature.

\citet{Arteta20163} uses as basis the method proposed in \cite{ISI:000371029500043}. The improvement is on the post-processing step, where nuclei region are refined using a surface. Two nuclei region has their optimal area defined by a smoothness factor.

Nuclei segmentation is proposed in \cite{ISI:000414283200217} based on an adaptive neighborhood provided by a regression tree. A comparison showed an improvement of 9\% with a nuclei segmentation without adaptive thresholding.

\begin{table*}[!ht]
\centering
\footnotesize
\begin{tabular}{l c l l} 
 \hline
 Reference & Year & Tissue region & Classifier \\
 \hline
 \hline
    \citet{Yu2008635} & 2008 & Gastric & SHMM \\
 \hline
    \citet{ISI:000371029500043} & 2012 & Breast & Structured SVM  \\
 \hline
 	\citet{Janssens20131206} & 2013 & Muscle & SVM \\
 \hline
 	\citet{Saraswat201444} & 2014 & Skin & \parbox[2cm]{3.5cm}{\rule{0pt}{1em}NSGA-II and \\Threshold classifier \rule[-0.3em]{0pt}{1em}} \\
 \hline
 	\citet{6999158} & 2014 & Breast & SVM \\
 \hline
 	\citet{Salman2014295} & 2014 & Prostate & KNN \\
 \hline
 	\citet{Chen2015} & 2015 & Breast & SVM \\
 \hline
 	\citet{ISI:000380546000311} & 2015 & \parbox[2cm]{3.5cm}{\rule{0pt}{1em}Epithelium \\(multiple tissues) \rule[-0.3em]{0pt}{1em}}& Support Vector Shape Segmentation \\
 \hline
	\citet{Wang20161} & 2016	& Breast & GA + SVM \\
 \hline
 	\citet{Arteta20163} & 2016 & Breast & Structured SVM \\
 \hline
 	\citet{Peikari20171078} & 2017 & Breast & SVM \\
 \hline
   	\citet{ISI:000414283200217} & 2017 & Regression Tree & Haar, fast radial symmetry \\
 \hline
\end{tabular}
\caption{Summary of publications focused on segmentation}
\label{table:segmentation_table}
\end{table*}
 	
%%%%%%%%%%%%%%%%%%%%%%%%%%%%%%%%%%%%%    
    
\subsubsection{Feature extraction}

Supervised shallow methods depend on the feature extraction of the data to provide a classification result. Some problems have their data already in a format that classifiers can interpret (e.g. data from sensors and laboratory exam results) and do not need a feature extractor. HI problems otherwise depend on a conversion from the images to a reasonable quantity numbers. Images itself are a matrix of numbers, but the number of values (attributes) for each instance is too high. As an example, images from BreaKHis \citep{Spanhol2016} dataset may generate a feature vector of 322 thousand values for each instance using only grayscale converted images. This is where feature extractors act. They process images and provide a reasonable number of attributes. Their role is not only dimension reduction, but more importantly, they can extract relevant information related to the problem (number of a determined element, textures, histograms) providing, if possible, a scale and rotation independent representation for example. Table \ref{table:feature_extraction_table} summarizes the articles related to feature extraction.

The search accomplished in this review found 14 articles focused on feature extraction placed inside the Supervised Shallow Methods subsection due to its close relation to this methods. Unsupervised shallow methods that deal with images in general also rely on feature extraction, but the works found usually provides a comparative result of the feature extraction method using shallow classifiers because results can be proved compared to the labeled data.

\citet{Ballaro2008703} proposes a segmentation of HI to identify sick or healthy megakaryocytes followed by feature extraction and classification. The most important fact highlighted in this article is the extraction of morphometric information for the classification. The classification is carried out by KNN and a regression tree.

\citet{Kuse2010235} uses a feature extraction procedure in a pre-segmentation process based on an unsupervised Mean-Shift Clustering. It reduces color variety to segment the image using thresholds. After this process nuclei are identified and have the overlapping removed by a contour and area restrictions. Finally, texture features based on GLCM are extracted using the segmented image and classified by an SVM classifier. The key aspect is the morphological restriction based on nuclei size. 

\citet{Huang2011579} proposes a classification schema based on a two-step feature extraction composed by a receptive field for detecting regions of interest and a sparse coding. The sparse coding groups feature from patches of the same region. The mean and covariance matrix of receptive field and sparse are the input of an SVM classifier.

\citet{Madabhushi2011506} highlights features that can be used to identify prostate cancer and provide prognosis based on the Gleason scale. The challenge of the work is how to fuse laboratory results with data from images (e.g. graph-based features) to create a meta-classifier merging multiple SVM classifiers.

\citet{CruzRoa201191} proposes a patching method on slides to create small regions and extract features from them to create a Bag of Words. Features from patches are obtained using SIFT, luminance level, and Discrete Cosine Transform. The classification process follows a multi-instance learning method based on the words represented by the described features. The underlying classifier is an SVM.

\citet{Caicedo2011519} combine seven feature extraction methods and create a kernel based representation of the data on each feature type. Kernels are used inside an SVM classifier to find similarity between data and implement a content retrieval mechanism.

\citet{Petushi2011} applies a segmentation process with Otsu thresholding to highlight nuclei. Nuclei provide the possibility to extract histograms of the following features: inside radial contact, inside line contact, area, perimeter, area vs. perimeter, curvature, aspect ratio and major axis alignment. These histograms are classified using a KNN classifier.

\citet{Loeffler20121867} uses inverse compactness and inverse solidness as measures for gland alteration on prostate cancer. They are obtained based on the area (object and convex hull area) and perimeter of threshold highlighted objects. These measures are used on a logistic regression to provide a relation to Gleason level.

\citet{Atupelage201361} proposes the creation of a feature extractor based on fractal geometry analysis. An SVM classifier was used to compare the proposed method with Gabor-filters, LM-Filters, LBP and GLCM features. The proposal outperformed the other methods achieving 95\% against 91\%, 73\%, 79\%, 86\% respectively.

Three morphological features are proposed by \citet{Song20131}, they are cystic cytoplasm length, cystic mucin production, and cystic cell density. These metrics are obtained after a segmentation by thresholding and watershed. SVM, KNN, Neural network and Bayesian classifiers evaluated the proposed features. The comparison is carried out using other morphological existing features. The results showed that the proposed three features attribute vector outperform the other 14 wide features vector achieving 90\% against 64\%. Interestingly combining all the 17 features the result is reduced to 85\%. 

\citet{FernandezCarrobles201525} presents a feature extraction method based on frequency and spacial textons. The use of textons implies that images are represented by a reduced vocabulary of textures. Features used for the classification are the histograms of textons and GLCM features of the image represented as a texton map. The study evaluates also the impact of different colormaps on these procedures. Six classifiers are used in the tests, SVM, Bagging, Fisher, Adaboost, LDA and Random Forests. The best results were 98.1\% with AdaBoost and Fisher classifier combining six color models and using GLCM for textons.

Features may be affected by the quality or even some minor variations during the process of acquisition and staining \cite{ISI:000391124500024}. An analysis using classification with Random Forest classifiers showed that the GLCM feature extractor is susceptible to image variations. GLCM is broadly used on HI classification works. Another point highlighted by the article is the importance of color normalization.

\citet{Noroozi2016128} proposes a feature extraction method based on Z-transforms and uses an SVM classifier to demonstrate the results. The proposed method outperformed three deep-learning feature methods achieving 85\% against 77\%, 77\% and 74\% on TICA, RICA, and SAE.

Random Forests and SVM are used in \cite{Fukuma20161202} to compare the feature extraction method using Voronoi Tessellation, Delaunay Triangulation and Minimum Spanning Tree (spatial-level descriptors) and Elliptical, Convex Hull, Bounding Box and Boundary (object-level descriptors). Object level extractors reached 99.07\% of accuracy at best and spatial ones obtained 82.88\%.

\citet{ISI:000403573100015} proposes Geometric aware features, based on Hu's moment, and Texture aware features, based fractal dimension, to detect texture and geometric changes in nuclei to differentiate between mitotic and non-mitotic cells. Random Forests were used to test the features' efficiency.

An LBP designed for multispectral images is proposed by \citet{Peyret201883}. It uses an SVM classifier to provide test the proposed feature extractor. The proposed LBP aligns all spectra and uses pixels from all other bands to calculate the LBP. It also uses a multi-scale kernel size. This feature extractor reaches 99\% of accuracy against 88.3\% comparing to standard LBP and 95.8\% of concatenated spectra LBP.

The classification between centroblast and non-centroblasts is the subject in \cite{Michail20143374}. It uses SVM, Decision Tree, Naive Bayes, KNN and QDA to study the morphometric, textural and color features in this classification context.

\citet{Olgun20141390} introduces a feature extractor for HI images called Local Object Pattern, based on the local distributions of objects. It measures the distance between an object and its neighborhood. A segmentation by color intensity before the feature extraction allows the identification of the objects. This method was tested using an SVM classifier and compared with another 13 methods between textural and structural. It outperforms all methods achieving 93\% against 90.3\% of the resampling-based Markovian model (RMM) (textural) and 92.2\% of a hybrid structural method.

\citet{Ozolek2014772} proposes the classification of follicular lesions on thyroid tissue using a preprocessing nuclei segmentation based on KNN. The feature extractor with Linear Optimal Transport provides data for the classification. An LDA evaluates the feature extraction method, that shows 100\% of agreement with all classifications except for the two classes for distinguishing between a follicular variant of papillary carcinoma and follicular carcinoma.

\citet{TambascoBruno2016329} uses a Curvelet Transform to handle multi-scale HI. LBP extracts features from Curvelets coefficients. An ANOVA analysis reduces the number of attributes of LBP features. This feature extraction method was tested using Decision Trees, SVM, Random Forests and Polynomial classifier. The best result was 100\% in comparison to other methods, where the best was 98.59\%.

A CAD system for bladder cancer is presented in \cite{KhalidKhanNiazi2016}. The focus is the extraction of epithelium features with segmentation using an automatic color deconvolution matrix construction. The accuracy performance was 88.2\%. 

\citet{ISI:000381691000001} investigates the best features for characterizing Lung cancer. They extracted 9,879 features using the Cell Profiler and tested using Naive Bayes, linear, polynomial and Gaussian SVM, Bagging of Trees, Random Forest and Breiman's Random Forest. A categorization separated features and information gain theory selected the most representative ones.

Fractal dimension is the feature set used in \cite{ISI:000391731800013} for breast cancer detection. It shows that these features perform well for a magnification of 40x to distinguish between malign and benign tumors, getting 97.9\% of F1-measure. On multiclass problem, it reaches only 55.6\%. All evaluations used SVM.

A mitosis detection system for breast cancer detection is presented by \cite{Wan2017291}. It uses the Dual-Tree Complex Wavelet Transform (DT-CWT) to represent the images. Generalized Gaussian Distribution and Symmetric alpha-Stable distributions parameters were the features for classification with SVM. The proposed method achieved 73\% of F-measure outperforming all compared methods except for one deep method, but the authors claimed less computational.

A feature extraction method based on nuclei textures is proposed in \cite{7532841} to improve classification performance. The proposed algorithm, called Adaptative Nuclei Shape Modeling, uses adaptive and iterative thresholding that considers nuclei area. The classification utilized Histograms of Oriented Gradients and LBP extracted from segmented nuclei. The proposed method achieved 93.3\% of accuracy against 92.3\%.

A different study from the majority of HI classification is presented in \cite{Reis20172344}. It uses the stroma maturity to evaluate breast cancer, but usually, the focus for identify cancer are nuclei. The features for the stroma are Basic Image Features (BIF), obtained by convolving images with a bank of Derivatives-of-Gaussian filters, and LBP with multiple scales of the neighborhood. RF classifier evaluates features. Proposed features achieved an accuracy of 80\% against 61\% of regular histograms.

Semantic features are high-level information that can be associated with an HI to aid its classification. \citet{ISI:000256869500006} proposes a low-level to high-level mapping to facilitate imaging retrieval. It consists of a gray histogram, color histogram, local binary partition, Tamura texture histogram, Sobel histogram, invariant feature histograms features. The results of 18 SVMs indicates what features better represent an image. The best feature is the information for comparison in the retrieval process.

\begin{table*}[!ht]
\centering
\footnotesize
\begin{tabular}{l c l l} 
 \hline
 Reference & Year & Tissue region & Feature \\
 \hline
 \hline
\citet{ISI:000256869500006} & 2008 & Skin & Semantic features (multiple SVM) \\
\hline
\citet{Ballaro2008703} & 2008 & Bone & Morphometric information \\
\hline
\citet{Kuse2010235} & 2010 & Lymphocyte & GLCM based \\
\hline
\citet{Huang2011579} & 2011 & Breast & Receptive field and sparse coding \\
\hline
\citet{Madabhushi2011506} & 2011 & Prostate & Multi-modal \\
\hline
\citet{CruzRoa201191} & 2011 & Skin & SIFT, Luminance, DCT \\
\hline
\citet{Caicedo2011519} & 2011 & Skin & \parbox[2cm]{4cm}{\rule{0pt}{1em}Histograms (grey and color)\\ Invariant feature histogram\\ Local binary patterns\\ SIFT \\Sobel histogram\\ Tamura texture histogram \rule[-0.3em]{0pt}{1em}}\\
\hline
\citet{Petushi2011} & 2011 & Breast & Nuclei shape \\
\hline
\citet{Loeffler20121867}  & 2012 & Prostate & Morphometric information \\
\hline
\citet{Atupelage201361} & 2013 & Blood & Fractal features  \\
\hline
\citet{Song20131} & 2013 & Pancreas & Morphometric information \\
\hline
\citet{FernandezCarrobles201525} & 2015 & Breast & Textons \\
\hline
\citet{ISI:000391124500024} & 2016 & Prostate & \parbox[2cm]{4cm}{\rule{0pt}{1em}Graph and subgraph based\\ Shape\\ Gland disorder \\ Texture \rule[-0.3em]{0pt}{1em}}\\
\hline
\citet{Noroozi2016128} & 2016 & Skin & Z-Transform \\
\hline
\citet{Fukuma20161202} & 2016 & Brain & Object and Spatial level features \\
\hline
\citet{Michail20143374} & 2014 & Lymphoma & Morphometric and textural \\
\hline
\citet{Olgun20141390} & 2014 & Colorectal & Local Object Pattern \\
\hline
\citet{Ozolek2014772} & 2014 & Thyroid & Linear Optimal Transport \\
\hline
\citet{TambascoBruno2016329} & 2016 & Breast & Curvelet Transform + LBP \\
\hline
\citet{KhalidKhanNiazi2016} & 2016 & Bladder & Morphometric \\
\hline
\citet{ISI:000381691000001} & 2016 & Lung & Quantitative and textural \\
\hline
\citet{ISI:000391731800013} & 2016 & Breast & Fractal Dimension \\
\hline
\citet{7532841} & 2016 & Uterus & Nucleus HOG and LBP \\
\hline
\citet{Reis20172344} & 2017 & Breast & Stroma features (LBP and BIF) \\
\hline
\citet{Wan2017291} & 2017 & Breast & \parbox[2cm]{4.2cm}{\rule{0pt}{1em}Wavelet transform\\ Gaussian distribution (GGD)\\Symmetric alpha-Stable \rule[-0.3em]{0pt}{1em}}\\
\hline
\citet{ISI:000403573100015} & 2017 & Oral & \parbox[2cm]{3.5cm}{\rule{0pt}{1em}Hu's moment\\ fractal dimension \\entropy \rule[-0.3em]{0pt}{1em}}\\
\hline
\citet{Peyret201883} & 2018 & Prostate & LBP \\
\hline
\end{tabular}
\caption{Summary of publications focused on feature extraction}
\label{table:feature_extraction_table}
\end{table*}

\normalsize

%%%%%%%%%%%%%%%%%%%%%%%%%%%%%%%%%%%%%%%%%%%%%%%%%%%%%%%%%%%%%%%%%%%%%%%%%%

\subsection{Deep methods}

The oldest work found \cite{Malon201297} uses and unspecified CNN to extract features from an image segmented with SVR and an SVM to classify the deep features. The purpose of the classification is to find mitotic nuclei. The remarkable aspect of the work is the comparison between the machine and three pathologists. The pathologists showed a Cohen Kappa factor of 0.13 and 0.44 in the best case, emphasizing the interobserver problem.

\citet{ISI:000337288300001} uses a custom CNN to classify patches of a WSI as invasive ductal carcinoma (breast cancer) or not. Patches end up labeled due to the region labeling. Some regions of the WSI, background and adipose cells, were excluded manually and were not patched. Patches were pre-processed using color normalization and the YUV colorspace. CNN outperformed the best handcraft feature extractor (Fuzzy Color Histogram) with a Random Forest classifier by 4\%. Compared to the other works analyzed, this one has a simple protocol and uses a small network, but it was a precursor or CNNs on HI.

In \cite{7950668} a VGG \cite{vgg2015} CNN performed a double segmentation on HI. A Random Forest classified the segmented images using Delaunay Triangulation and Area-Voronoi diagram features. This work focused on stroma characteristics. Another concept applied in this article is the use of data augmentation. In \cite{ISI:000337288300001}, data augmentation was not necessary because it considered patches, which intrinsically increased the number of images. Both works were related to breast cancer. Another work that involves breast cancer and CNN is \cite{7899848}, which uses AlexNet and compares the results with some hand-crafted feature extractors. The deep learning approach did not outperform the shallow methods. Breast cancer was also the subject in \cite{ISI:000406771302072}, where a multi-task CNN was trained to identify malignancy of tumors and magnification factors of HI. Data augmentation was used as in \cite{7950668} with rotation and flipping transformations.

Mitosis is the matter of the works presented by \citet{ISI:000375550500015} and \citet{7950667}. Both works used breast cancer-based datasets. The first work proposes a high-level framework classify labeled data from specialists and crowdsourcing. The idea is to test the resilience of CNNs to noisy information that comes from non-specialists. The second work presents the results of a Wide Residual Network on mitotic recognition. It outperformed most of the compared results in the field, but it states that pre-processing can improve the results.

In \cite{Khosravi2018317} the versatility of CNNs was tested using eight different datasets (breast, lung, and bladder tissues) with H\&E and immunohistochemistry (IHC) images. The tests included Inception, Resnet, a combination of them and the transfer learning concept. Results showed good performance in spite of the raw images, with no pre-processing procedure. 

The colorectal cancer is the subject of five deep-learning works. All of them are between 2016 and 2017. Two works, \cite{8037745} and \cite{7950492} used the same dataset, published in \cite{Kather2016} (CRC). The first of them employ a Bilinear Convolutional Neural Network (BCNN), that consists of two CNNs with a bilinear function at the end to combine the results. The input data of the CNNs are images with the hematoxylin and eosin channels separated in a prior stage by a color decomposition algorithm. The results are not the classification provided by the BCNN but predicted by an SVM classifier with the BCNN as feature extractor. The second work related to CRC dataset applies a custom CNN to analyze the image preprocessing, corroborating to the affirmations of some works found in this survey about preprocessing.

Three deep learning works used a dataset \cite{7109172} of gland segmentation. \citet{7493530} compares both hand-crafted features, CNN extracted features and CNNs. A combination of hand-crafted features classified with an SVM and the prediction of a CNN showed the best results. This work used data augmentation with rotations and mirroring for hand-crafted features and CNNs. The work presented by \citet{7885586} segments and distinguishes glands. It presents an approach using data augmentation and multiple CNNs to perform background separation, object detection, and edge detection. The results of these three networks feed another CNN that outputs a segmented image. The last work \cite{Kainz2017} applied color deconvolution to preprocess images and multiple CNNs to the classification. One of the CNNs identifies four types of structures, and a probabilistic merge combines its result with the ones of a CNN specialized in separate foreground and background. 

In \cite{ISI:000411791700059} CNNs are compared to shallow classifiers (SVM, Random Forest, KNN and Naive Bayes). CNNs outperformed the hand-crafted features with shallow classifiers to identify prostate cancer. 
The pre-processing of input images included color deconvolution to unmix the stains and data augmentation with affine transforms.

A multi-modal approach with two-step classification is the subject in \citet{Carneiro20171405}.  A image segmentation with five classifiers, including a CNN precedes the CNN classification step. The approach used simultaneously two types of images for segmentation, H\&E, and immunofluorescence. Besides the use of a CNN, the work proposes the FLSSVM which outperforms the CNN.

A comparison between unsupervised and supervised feature learning in brain tumors is presented in \cite{ISI:000349328300013}. The conclusions are focused on the network depth and on what characteristics of HI CNNs extract.
    
A work with a dataset of gastric tumors is presented in \cite{ISI:000416616500002}. It proposes a custom CNN and compares it to an AlexNet and traditional feature extractors and classifiers. As in other works, affine transforms generate data augmentation. The custom CNN outperformed other compared methods only for necrosis detection, not for tumor detection.
    
In \cite{ISI:000366205700077} it is presented a proposal of feature extraction using Deconvolution Networks.

The work presented by \citet{Oikawa2017} used a hybrid approach to classify benign and malign tissues from gastric cancer. The last step in their work is the classification using a CNN in case of the previous SVM classifier could not distinguish the type of tumor.

\citet{Malon201297} compared the agreement of three pathologists and a machine learning method. An SVM classifies images' feature vectors built with the merging of deep, shape, color, mass and texture features. Cohen Kappa between pathologists were 0.13, 0.44 and 0.39. The results of accuracy for ML method were 63.6\% and 98.6\% for positive and negative cases respectively, close to two pathologists' performance. Only one pathologist performed 99.2\% and 94.5\% on positive and negative samples.

	Deep learning approaches have been outperforming traditional classifiers in recent years, although studies are still necessary for understanding how these networks learn, in special in the context of HI. HI present different characteristics than images of our macroscopic world, like the shapes of objects and letters.

%%%%%%%%%%%%%%%%%%%%%%%%%%%%%%%%%%%%%%%%%%%%%%%%%%%%%%%%%%%%%%%%%%%%%%%%%%

\subsection{Reviews}

This section brings a summary in Table \ref{table:reviews_table} of the reviews related to histopathology images and machine learning methods. We found 18 works in this category. \citet{Saha2016461}, \citet{Nawaz2016296}, \citet{Chen2017} and \citet{Robertson2017} are works published in medical journals, so they provide superficial comments towards Machine Learning methods, but they provide a deeper view of the histology information. The review of \citet{Nawaz2016296} is an example of a deep analysis of the characteristics of tumors and a brief study of how computational methods can deal with this problem.

The work presented in \cite{Komura2018} presents how Machine Learning methods can be employed in HI, it also presents a series of datasets from this field. \citet{Litjens201760} reviews Deep Learning methods applied to a variety of medical images. Reviews between 2012 and 2015 are more concerned about nuclei segmentation and classification. After 2015 works focused on the classification of the whole images.

\begin{table*}[!ht]
\centering
\scriptsize
\begin{tabular}{l l l l} 
 \hline
 Reference & Type of images & Characteristics of the review & Journal or Conference \\
 \hline
 \hline
 \citet{He2012538} & HI &\parbox[c][0.7cm][c]{3.5cm}{\setlength{\baselineskip}{1.0pt} Segmentation, Feature \\ extraction and classification}& Computer Methods and Programs in Biomedicine \\
 \hline
 \citet{6690201}  & HI and IHC & \parbox[2cm]{4.0cm}{\setlength{\baselineskip}{1.0pt}\rule{0pt}{1em}Nuclei extraction, Segmentation\\Feature extraction and classification \rule[-0.3em]{0pt}{1em}}& IEEE Reviews in Biomedical Engineering \\
 \hline
 \cite{6745404} & HI, IHC and other specifics & Segmentation &  \parbox[2cm]{4.8cm}{\setlength{\baselineskip}{1.0pt}\rule{0pt}{1em}2014 International Conference on Electronic\\Systems, Signal Processing and Computing Technologies} \\
 \hline
 \citet{7192913} & HI & Nuclei segmentation and classification &  \parbox[2cm]{4.8cm}{\setlength{\baselineskip}{1.0pt}\rule{0pt}{1em}2015 International Conference on Innovations in Information, Embedded and Communication Systems\rule[-0.3em]{0pt}{1em}} \\
 \hline
 \citet{Veta2015237} & HI & Results from MITOS 2013 challenge & Medical Image Analysis \\
 \hline
 \citet{Nawaz2016296} & Various & Tumor ecology & Cancer Letters \\
 \hline
 \citet{Madabhushi2016170} & HI & \parbox[2cm]{3.5cm}{\setlength{\baselineskip}{1.0pt}\rule{0pt}{1em}Detection, segmentation\\feature extraction and classification\rule[-0.3em]{0pt}{1em}} & Medical Image Analysis \\
 \hline
 \citet{Saha2016461} & HI & \parbox[0.9cm]{5.3cm}{\setlength{\baselineskip}{1.0pt}Slide preparation, staining, microscopic imaging,\\ pre-processing, segmentation, feature extraction \\ and diagnostic classification for breast cancer} & Tissue and Cell \\
 \hline
 \citet{Chen2017} & HI & Image analysis of H\&E slides for breast cancer prognosis & Tumor Biology \\
 \hline
 \citet{Robertson2017} & Various & Deep Learning and breast cancer & Translational Research \\
 \hline
 \citet{Cosma20171} & \parbox[2cm]{2.2cm}{\setlength{\baselineskip}{1.0pt}\rule{0pt}{1em}HI \\ Other type of images \rule[-0.3em]{0pt}{1em}} & \parbox[2cm]{6cm}{\setlength{\baselineskip}{1.0pt}\rule{0pt}{1em}Deep and shallow methods\\ for prostate cancer detection \rule[-0.3em]{0pt}{1em}}& Expert Systems with Applications \\
 \hline
 \citet{AzevedoTosta201735} & HI & Segmentation for lymphocytes & Informatics in Medicine Unlocked \\ 
 \hline
 \citet{Litjens201760} & Medical Images & Deep learning for all types of medical images & Medical Image Analysis \\
 \hline
 \citet{DiCataldo201756} & HI & Feature extraction & Computational and Structural Biotechnology Journal\\
 \hline
 \citet{Aswathy201774} & HI & Image processing and classification for breast cancer & Informatics in Medicine Unlocked \\
 \hline
 \citet{Li201866} & Medical images & Content retrieval & Medical Image Analysis \\
 \hline
 \citet{Komura2018} & HI & \parbox[1cm]{4.3cm}{\setlength{\baselineskip}{1.0pt} \rule[-0.3em]{0pt}{1em}Datasets and machine learning methods\\for multiple problems} & Computational and Structural Biotechnology Journal \\
 \hline

\end{tabular}
\caption{Summary of the reviews about HI and Machine Learning}
\label{table:reviews_table}
\end{table*}

%%%%%%%%%%%%%%%%%%%%%%%%%%%%%%%%%%%%%%%%%%%%%%%%%%%%%%%%%%%%%%%%%%%%%%%%%%

\normalsize

\section{Case study}

In this section, we present results for the classification of the BreaKHis dataset in two classes, malign and benign. We used a patching procedure to obtain a set of sub-images from the original ones. The idea is to improve the quality of the images provided to the classifier. We used an SVM classifier with RBF kernel and two feature extractors: PFTAS and Deep features. We proposed a filtering technique in order to remove irrelevant patches that could be considered irrelevant to the classification process. This section is subdivided into Datasets, which presents the datasets used in the process. BreaKHis patches filtering describes how we created the patches' filter. In the subsection "Results of filtering with PFTAS", we describe and comment the first results with the handcrafted feature extractor. In subsection 3.4 we explain the deep features extraction. The subsection 3.5 presents the results with deep features. 

\label{casestudy}

\subsection{Datasets}

\subsection*{BreaKHis}
The study case was accomplished on a dataset of breast cancer histopathologic images called BreaKHis. The dataset, published by \cite{Spanhol2016}, contains 7909 images of 82 patients distributed in eight classes of tumors. Four of the eight classes are malign tumors and other four, benign. The dataset is imbalanced by a factor of seven in the worst case, which means e.g. that ductal carcinoma (malign) images have seven times more samples than adenosis (benign). Table \ref{table:classdistribution} shows the class distribution.
\begin{table}[ht!]
\centering
\footnotesize
\begin{tabular}{l l r r} 
 \hline
 & Tumor type & Images & Patients \\
 \hline
\multirow{5}{*}{{\rotatebox[origin=c]{90}{\parbox[c]{1cm}{\centering Benign}}}} & Adenosis & 444 & 4 \\
& Fibroadenoma & 1014 & 10 \\
& Tubular adenoma & 453 & 3 \\
& Phyllodes tumor & 569 & 7 \\
& Total & 2368 & 24 \\
\hline
\multirow{5}{*}{{\rotatebox[origin=c]{90}{\parbox[c]{1cm}{\centering Malign}}}} & Ductal carcinoma & 3451 & 38 \\
& Lobular carcinoma & 626 & 5 \\
& Mucinous carcinoma & 792 & 9 \\
& Papillary carcinoma & 560 & 6 \\
& Total & 5429 & 58 \\
 \hline
\end{tabular}
\caption{Image and patient distribution of BreaKHis dataset}
\label{table:classdistribution}
\end{table}
Each image has 700x460 pixels. All patients contain images with four magnification factors, 40x, 100x, 200x e 400x equivalent to 0.49, 0.20, 0.10, 0.05 $\mu$m per pixel. Images of different magnification were obtained from the minor to the major magnification factor when the pathologist selects a region of interest to be analyzed more carefully enlarging it with a more powerful lens.

\subsection*{Colorectal Cancer (CRC)}

The dataset presented in \cite{Kather2016}, referred now on as CRC, is a dataset of colorectal cancer histopathology images of 5000 x 5000 pixels that were patched into 150x150 pixels images and labeled according to the structure they contain. Eight types of structures are labeled: Tumor (T), Stroma (ST), Complex Stroma (C), Immune or lymphoid cells (L),  Debris (D), Mucosa (M), Adipose (AD), and Background or empty(E). The total number of images is 625 per structure type, resulting in 5000 images.

\subsection{BreaKHis patches filtering}

We developed a method for classifying BreaKHis dataset images into two classes, malign and benign. Our approach followed the same protocol used in the work that proposed the dataset. The protocol uses 5-fold cross-validation with predefined folds. The split percentage is patient-wise with around 70\% of patients for training and 30\% for testing. This procedure warranties that all images of a patient are on the test set or in training set, but not in both at the same experiment.

\begin{figure}
	\centering
    \includegraphics[width=3.4in]{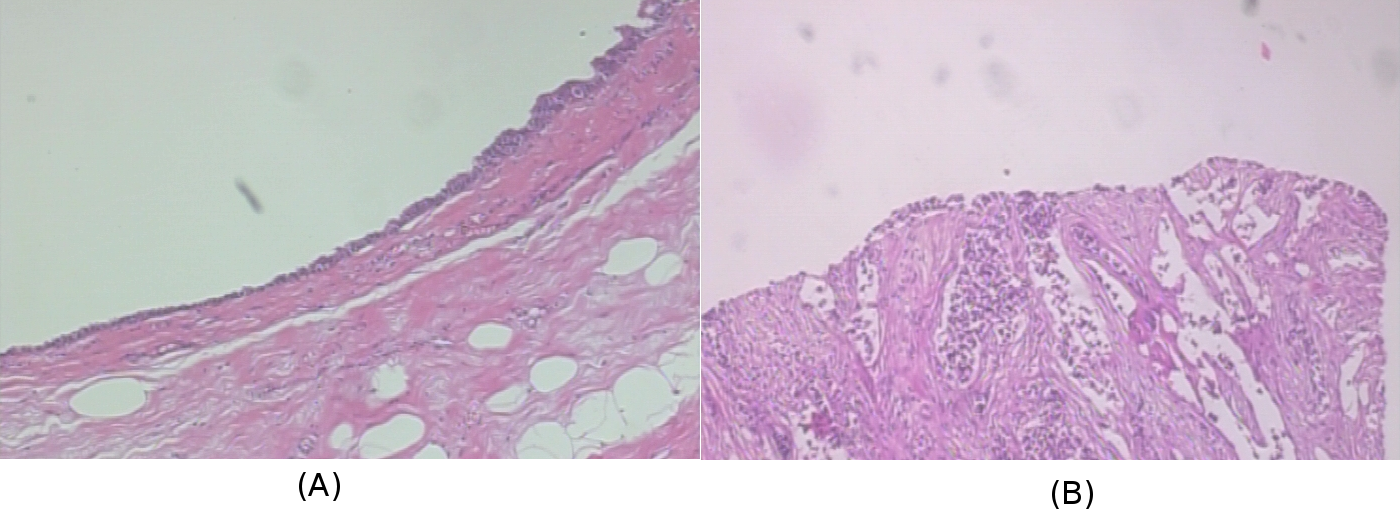}
	\caption{A - Example of a Adenosis (benign tumor) slide, B - Example of a Ductal Carcinoma (malign tumor) slide}
    \label{fig:adenoma_ductal_carcinoma}
\end{figure}

Our approach relies on the fact that some images of different classes from this dataset contain lots of information in common. Figure \ref{fig:adenoma_ductal_carcinoma} depicts an example of two opposite types of tumor, an Adenoma (benign) and a Ductal Carcinoma (malign) that have a similar appearance in some area. The similar content is the background of the slide. We propose patching the images and exclude patches considered irrelevant to the classification. We consider no prior knowledge of tissue types inside each image this means that the content inside each image is not separated and labeled. The labeling only tells what type of tumor an entire image contains.

\begin{figure}
	\centering
    \includegraphics[width=3.6in]{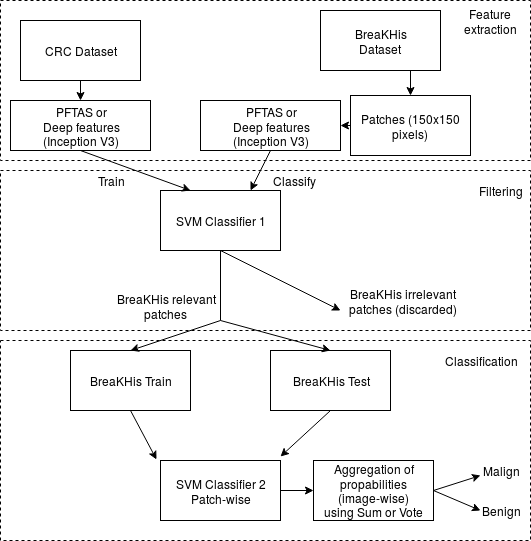}
	\caption{Filtering process}
    \label{fig:filtering}
\end{figure}

The main idea is to transfer the knowledge contained in the CRC dataset to the BreaKHis dataset eliminating patches that do not contain relevant information or even similar patches of distinct classes. The patch elimination process is called filter and is illustrated in Figure \ref{fig:filtering}. The filtering tool is an SVM classifier trained with CRC's images. Our purpose is not to classify all the structures present in BreaKHis' images, but only to eliminate what we call irrelevant. The key to the filtering process was to define what is relevant and what is irrelevant. Intuitively only background is considered irrelevant, but we tested the combination of some structures. Table \ref{table:structurefilter} shows how the number of images of each class of CRC dataset was distributed to create the filter with only two classes. 
We proposed to reduce the amount of CRC images in the extreme scenarios of filters 1 and 7 to reduce the imbalance. In these scenarios, we have the same total number of images to relevant and irrelevant. The quantity of images of each sub-class inside each class is the same (e.g. 89 images of ST, C, L, D, M, A, E inside irrelevant class for filter 1), but the ratio between sub-classes of irrelevant and relevant are imbalanced  (e.g. 89 ST for irrelevant and 625 T for relevant for filter 1). This problem is worse in filters 1 and 7, not so bad for filters 2 and 6, better for 3 and 5 and does not occur in filter 4. This decision aims to not create a bias inside the filter between relevant and irrelevant classes. 

\begin{table}[ht!]
\footnotesize
\centering
\begin{tabular}{|c |c | c | c| c| c| c| c| c|} 
 \hline
	 &T & ST & C & L & D & M & A & E \\
     &625 & 625 & 625 & 625 & 625 & 625 & 625 & 625 \\
	\hline
     \hline
	\multirow{3}{*}{1} &\multicolumn{1}{|c|}{Relevant} & \multicolumn{7}{c|}{Irrelevant} \\
	&\multicolumn{1}{|c|}{625 images} & \multicolumn{7}{c|}{625 images} \\ \cline{2-9}
    & 625 & 89 & 89 & 89 & 89 & 89 & 89 & 89 \\
	\hline
     \hline
	\multirow{3}{*}{2} & \multicolumn{2}{|c|}{Relevant} & \multicolumn{6}{c|}{Irrelevant} \\
	&\multicolumn{2}{|c|}{1250} & \multicolumn{6}{c|}{1248} \\ \cline{2-9}
	&625 & 625 & 208 & 208 & 208 & 208 & 208 & 208 \\
	\hline
     \hline
	 \multirow{3}{*}{3} & \multicolumn{3}{|c|}{Relevant} & \multicolumn{5}{c|}{Irrelevant} \\
	& \multicolumn{3}{|c|}{1875} & \multicolumn{5}{c|}{1875} \\
     \cline{2-9}
      & 625 & 625 & 625 & 375 & 375 & 375 & 375 & 375 \\
     \hline
	\hline
	 \multirow{3}{*}{4} & \multicolumn{4}{|c|}{Relevant} & \multicolumn{4}{c|}{Irrelevant} \\
	& \multicolumn{4}{|c|}{2500} & \multicolumn{4}{c|}{2500} \\
     \cline{2-9}
     & 625 & 625 & 625 & 625 & 625 & 625 & 625 & 625 \\
	\hline
	\hline
	\multirow{3}{*}{5} & \multicolumn{5}{|c|}{Relevant} & \multicolumn{3}{c|}{Irrelevant} \\
	& \multicolumn{5}{|c|}{1875} & \multicolumn{3}{c|}{1875} \\
     \cline{2-9}
      & 375 & 375 & 375 & 625 & 625 & 625 & 625 & 625 \\
     \hline
	\hline
	 \multirow{3}{*}{6} & \multicolumn{6}{|c|}{Relevant} & \multicolumn{2}{c|}{Irrelevant} \\
	& \multicolumn{6}{|c|}{1248} & \multicolumn{2}{c|}{1250} \\
     \cline{2-9}
	& 208 & 208 & 208 & 208 & 208 & 208 & 625 & 625 \\
     \hline
	\hline
	 \multirow{3}{*}{7} & \multicolumn{7}{|c|}{Relevant} & \multicolumn{1}{c|}{Irrelevant} \\
	& \multicolumn{7}{|c|}{625 images} & \multicolumn{1}{c|}{625 images} \\
     \cline{2-9}
     & 89 & 89 & 89 & 89 & 89 & 89 & 89 & 625 \\
     \hline     
\hline
\end{tabular}
\caption{Number of images of CRC dataset inside each new category (relevant and irrelevant)}
\label{table:structurefilter}
\end{table}
We trained the filter with 85\% of data and left 15\% of images to the validation set. No test set was necessary because the purpose was not to evaluate the filter but the filtering process. The selection between train and validation was random and stratified. Seven SVM classifiers were trained using grid search and 5-fold cross-validation on the data distribution shown in Table \ref{table:structurefilter}. Parameter-Free Threshold Adjacency Statistics (PFTAS), implemented by the library Mahotas \cite{mahotas2013}, was chosen to extract features resulting in a 162-dimensional feature vector. We used the same feature extractor for both CRC and BreaKHis images. Figure \ref{fig:filtering} shows feature extraction with PFTAS or Deep Features (InceptionV3). The next subsection explains the Deep Features method.
\par
BreaKHis' 7909 images generated 118635 patches of 150 x 150 pixels. This size is the same as the CRC images. Keeping the image pixel size close does not make images completely compatible, because pixels of BreaKHis' images have 49 $\mu$m and CRC images have 74 $\mu$m in the most compatible case. The pixel density problem cannot be solved because it is an acquisition characteristic. Patches of BreaKHis have a small vertical overlap of approximately five pixels and approximately 12 pixels horizontally caused by the distribution of 5 patches of 150 pixels in the 700 pixels wide image horizontally and in 460 pixels height.
\par
All patches were submitted to the same filtering process, independently on their magnification factor. Table \ref{table:filtering} shows the number of patches, images, and patients selected as relevant after filtering and separated by the magnification factor.

\begin{table}[ht!]
\centering
\footnotesize
\begin{tabular}{c r r r r} 
 \hline
 Magnif. & Filter & Patches & Images & Patients \\
 \hline
 \multirow{7}{*}{100} & 1 & 39.0 & 6.1 & 0.8 \\
 & 2 & 96.3 & 60.3 & 21.3 \\
 & 3 & 93.9 & 40.7 & 10.4 \\
 & 4 & 97.5 & 51.7 & 14.7 \\
 & 5 & 100.0 & 95.8 & 68.7 \\
 & 6 & 91.4 & 88.5 & 70.2 \\
 & 7 & 100.0 & 97.3 & 93.7 \\
 \hline
 \multirow{7}{*}{200} & 1 & 47.5 & 9.4 & 1.3 \\
 & 2 & 100.0 & 61.5 & 21.3 \\
 & 3 & 92.6 & 32.8 & 7.6 \\
 & 4 & 97.5 & 41.9 & 10.1 \\
 & 5 & 100.0 & 87.4 & 53.3 \\
 & 6 & 97.5 & 89.0 & 59.7 \\
 & 7 & 100.0 & 99.8 & 97.3\\
 \hline
 \multirow{7}{*}{400} & 1 & 76.8 & 22.5 & 4.6 \\
 & 2 & 98.7 & 66.7 & 19.7 \\
 & 3 & 90.2 & 34.5 & 6.4 \\
 & 4 & 95.1 & 33.7 & 5.8 \\
 & 5 & 100.0 & 82.3 & 43.3 \\
 & 6 & 100.0 & 89.8 & 47.7 \\
 & 7 & 100.0 & 100.0 & 99.6 \\
 \hline
 \multirow{7}{*}{40} & 1 & 48.7 & 10.6 & 1.7 \\
 & 2 & 100.0 & 73.0 & 31.5 \\
 & 3 & 95.1 & 62.7 & 24.1 \\
 & 4 & 98.7 & 75.4 & 33.5 \\
 & 5 & 100.0 & 99.5 & 81.3 \\
 & 6 & 93.9 & 88.4 & 70.1 \\
 & 7 & 100.0 & 96.9 & 90.9 \\
 \hline
\end{tabular}
\caption{Percentage of patches, images and patients considered relevant after the filtering process after each filter}
\label{table:filtering}
\end{table}

After the filtering process, we proceeded with the classification process, which used the filtered patches extracted from BreaKHis images classified as relevant. Filtered patches were divided into train and test sets exactly the same way as published on \cite{Spanhol2016}, so five folds were used. We obtained the best parameters for the SVM classifier with grid search.  We executed seven experiments, one for each filter. All filters were executed separately for each one of the four magnification factors. The five folds made necessary all the 28 executions of filters with magnification factors be repeated five times, resulting in 140 executions. We executed one more batch of experiments without the filter, adding five more folds for each four magnification factors adding more 20 experiments, these variations summed up to 160 experiments.

\subsection{Results of filtering with PFTAS}

Tables \ref{table:resultsnofiltering} and \ref{table:resultsfiltering} present the results of the first experiments in patch level, image level, and patient level. The two columns of results for image and patient-level are the sum and vote aggregation functions \citep{667881} used to combine the prediction probability results of patches into an image. The classification result of an image is the aggregation function of patches' results. The image classification result (from patch aggregation) is used to calculate the patient level results. 

The calculation of the accuracy per patient is obtained by Equation \ref{ref:eq1}, where $N_{correct}$ is the number of images correct classified of one patient and $N_{total}$ is the total images of a patient.

\begin{equation}
\label{ref:eq1}
Patient score = \frac{N_{correct}}{N_{total}}
\end{equation}

The overall accuracy is obtained by Equation \ref{ref:eq2}, where \textit{Patient score} comes from Equation \ref{ref:eq1} and \textit{Total number of patients} is the number of patients.

\begin{equation}
\label{ref:eq2}
Accuracy = \frac{\sum Patient score}{Total\ Number\ of\ Patients}
\end{equation}

\begin{table}[ht!]
\centering
\footnotesize
\setlength{\tabcolsep}{4.3pt}
\renewcommand{\arraystretch}{1.2}
\begin{tabular}{c c c c c c c} 
 \hline
 Magnif.& Patches & \multicolumn{2}{c}{Images} & \multicolumn{2}{c}{Patients} \\
 & & Sum & Vote & Sum & Vote \\
 \hline
40 & 82.9 $\pm$ 3.4 & 85.0 $\pm$ 4.4 & 85.4 $\pm$ 4.4 & 86.1 $\pm$ 4.5 & 86.4 $\pm$ 4.7 \\
100 & 83.0 $\pm$ 3.8 & 84.9 $\pm$ 3.9 & 84.6 $\pm$ 3.9 & 86.6 $\pm$ 4.9 & 86.3 $\pm$ 4.9 \\
200 & 86.5 $\pm$ 3.1 & 88.3 $\pm$ 3.5 & 88.5 $\pm$ 3.6 & 88.4 $\pm$ 4.6 & 88.7 $\pm$ 4.5 \\
400 & 84.8 $\pm$ 4.5 & 87.0 $\pm$ 4.8 & 87.3 $\pm$ 4.8 & 88.0 $\pm$ 5.7 & 88.2 $\pm$ 5.4 \\
\hline
\end{tabular}
\caption{Results for 4 magnification factors without filter}
\label{table:resultsnofiltering}
\end{table}

\begin{table}[ht!]
\centering
\footnotesize
\setlength{\tabcolsep}{4pt}
\renewcommand{\arraystretch}{1.2}
\scalebox{1.0}{
\begin{tabular}{c c c c c c c}
 \hline
Magnif.& Patches & \multicolumn{2}{c}{Images} & \multicolumn{2}{c}{Patients} \\
 & & Sum & Vote & Sum & Vote \\
 \hline
40 & 82.5 $\pm$ 3.3 & 85.1 $\pm$ 4.6 & 84.9 $\pm$ 5.2 & 86.4 $\pm$ 4.3 & 86.1 $\pm$ 4.9 \\
100 & 82.7 $\pm$ 3.6 & 85.4 $\pm$ 3.9 & 84.8 $\pm$ 3.6 & 87.0 $\pm$ 4.3 & 86.6 $\pm$ 4.1 \\
200 & \textbf{86.7} $\pm$ 3.6 & \textbf{89.0} $\pm$ 3.9 & \textbf{88.9} $\pm$ 4.3 & \textbf{89.2} $\pm$ 5.1 & \textbf{89.3} $\pm$ 5.3 \\
400 & 84.8 $\pm$ 4.6 & 87.0 $\pm$ 4.9 & 87.2 $\pm$ 5.0 & 87.9 $\pm$ 5.7 & 88.2 $\pm$ 5.8 \\
\hline
\end{tabular}}
\caption{Results for 4 magnification factors with the filter 7, considering only the Empty label of CRC as irrelevant}
\label{table:resultsfiltering}
\end{table}

We considered in Table \ref{table:resultsfiltering} only the results with the filter number seven, that considers as irrelevant only the Empty images of CRC dataset because it produced better results and never excluded patients. Only magnification factor 200 got improvements with filter application as highlighted. It is important to note that the best results presented in \cite{Spanhol2016}, with SVM classifier and PFTAS feature extractor, reached 81.6$\pm$3.0, 79.9$\pm$5.4, 85.1$\pm$3.1 and 82.3$\pm$3.8 for 40x, 100x, 200x and 400x respectively. Our proposal using only the patching process improved the results compared to the initial results.

\subsection{BreaKHis patches filtering using Deep Features}

We proposed another set of experiments using feature extraction based on Convolutional Neural Networks (CNN). We used an InceptionV3 \citep{SzegedyVISW15} to extract features from both, CRC and BreaKHis patches. The feature extractor was the last layer before the softwax fully connected layer that presents the predictions for the ImageNet classes. We used an Inception network pre-trained with ImageNet to generate a feature vector with 2048 elements per instance.

The intention of this feature extraction procedure was to verify if other features can improve the detection of irrelevant regions. We used the Principal Component Analysis (PCA) method to reduce the feature vector dimensionality from 2048 to 100, 200, 400 and 600 attributes. We chose four values to study the impact of the reduction. The value of 600 features corresponds to 95\% of the accumulated variance from the most important components.

\subsection{Results of filtering using Deep Features}

\begin{figure}
	\centering
    \includegraphics[width=3.2in]{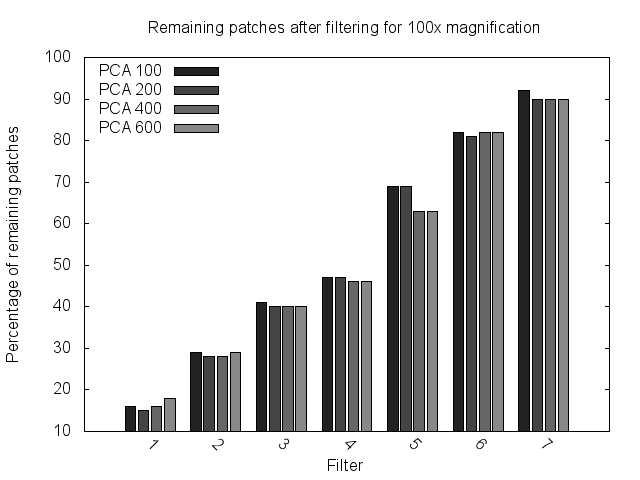}
	\caption{Remaining patches for each filter}
    \label{fig:filter_statistics_tensorflow}
\end{figure}

The results of the filtering process in terms of patch reduction are in Figure \ref{fig:filter_statistics_tensorflow}. This graph represents only the remaining patches for magnification factor 100x. All other magnifications (200x, 400x and 40x) follows the same distribution. Aggressive filters can exclude all images from a patient and in this case, we did not consider the results for these filters. Some filters excluded all patches from an image, we do not consider this a problem, meaning that the excluded image does not contribute to the classification.

% Figures \ref{fig:results_tensorflow_40}, \ref{fig:results_tensorflow_100}, \ref{fig:results_tensorflow_200} and \ref{fig:results_tensorflow_400} present results for classification using an SVM classifier and Deep Features. Four groups inside each graph corresponds to the sizes of feature vectors reduced with PCA.

% \begin{figure}
% 	\centering
%     \includegraphics[width=3.2in]{Figures/40x.png}
% 	\caption{Accuracy for patient level classification for 40x magnification with InceptionV3 extracted features}
%     \label{fig:results_tensorflow_40}
% \end{figure}

% \begin{figure}
% 	\centering
%     \includegraphics[width=3.2in]{Figures/100x.png}
% 	\caption{Accuracy for patient level classification for 100x magnification with InceptionV3 extracted features}
%     \label{fig:results_tensorflow_100}
% \end{figure}

% \begin{figure}
% 	\centering
%     \includegraphics[width=3.2in]{Figures/200x.png}
% 	\caption{Accuracy for patient level classification for 200x magnification with InceptionV3 extracted features}
%     \label{fig:results_tensorflow_200}
% \end{figure}

% \begin{figure}
% 	\centering
%     \includegraphics[width=3.2in]{Figures/400x.png}
% 	\caption{Accuracy for patient level classification for 400x magnification with InceptionV3 extracted features}
%     \label{fig:results_tensorflow_400}
% \end{figure}

\begin{table}[ht!]
\centering
\footnotesize
\setlength{\tabcolsep}{4pt}
\renewcommand{\arraystretch}{1.2}
\scalebox{1.0}
{

\begin{tabular}{cccrrrr}
Ft. Number & Mag. & Filter Nr. & \multicolumn{2}{c}{Filter}  & \multicolumn{2}{c}{No filter} \\
 & & & Mean & Std. Dev. & Mean & Std. Dev. \\
\hline
\multirow{4}{*}{PCA 100} & 40 & 6 & 89.9 & 3.6 & 88.5 & 3.8 \\
& 100 & 7 & 91.0 & 3.0 & 90.3 & 3.4 \\
& 200 & 7 & 89.7 & 3.6 & 88.6 & 3.6 \\
& 400 & 6 & 86.7 & 1.6 & 85.7 & 2.4 \\
\hline
\multirow{4}{*}{PCA 200} & 40 & 6 & 88.7 & 2.9 & \textbf{89.0} & 3.4 \\
& 100 & 7 & 90.6 & 3.4 & 90.0 & 3.7 \\
& 200 & 7 & 89.7 & 3.0 & 89.1 & 2.7 \\
& 400 & 7 & 86.9 & 2.2 & 85.4 & 2.6 \\
\hline
\multirow{4}{*}{PCA 400} & 40 & 6 & 89.6 & 3.4 & 89.2 & 3.6 \\
 & 100 & 7 & 90.1 & 3.7 & 89.8 & 3.8 \\
 & 200 & 7 & 89.7 & 2.9 & 89.1 & 3.0 \\
 & 400 & 6 & 87.1 & 1.5 & 85.3 & 1.7 \\
\hline
\multirow{4}{*}{PCA 600} & 40 & 6 & 89.5 & 3.4 & 89.3 & 3.5 \\
& 100 & 7 & 89.8 & 3.4 & 89.7 & 3.7 \\
& 200 & 7 & 89.7 & 3.2 & 89.1 & 2.9 \\
& 400 & 7 & 86.9 & 1.8 & 85.6 & 2.0 \\
\hline
\end{tabular}}
\caption{Accuracy for features extracted with InceptionV3 and classified with a SVM classifier, in bold the only scenario where data without filter outperformed filtered data}
\label{tab:results_tensorflow}
\end{table}

\begin{figure}[!ht]
	\centering
    \includegraphics[width=3.4in]{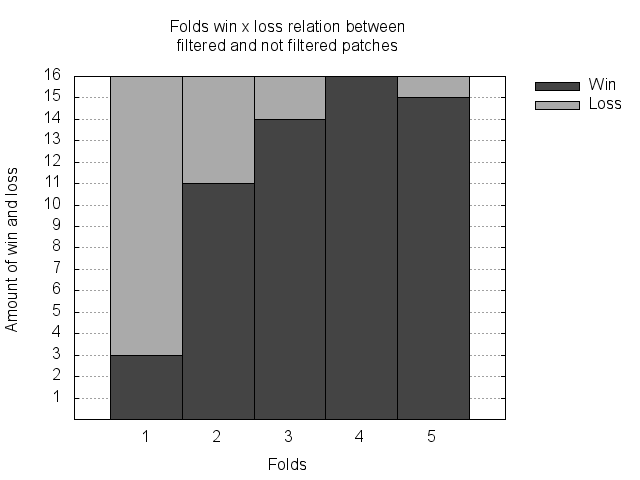}
	\caption{Comparing of losses and wins inside each folder for best filter and no filter execution}
    \label{fig:loss_win}
\end{figure}

Table \ref{tab:results_tensorflow} shows the results of graphs where it is possible to see that the filtering process only loses in the magnification factor of 40x and with feature vector size of 200 attributes (value in bold). Figure \ref{fig:loss_win} shows the relation between win and loss for the filtered and not filtered execution. It is possible to note that the filtering process produces worst results only on fold 1 of all executions comparing to the 16 results presented in Table \ref{tab:results_tensorflow}. For fold 1, it loses 13 times, but for the other folds, there are more wins than losses.

\section{Conclusion}
\label{conclusion}
In this study, we presented a literature review of the Machine Learning methods usually employed in histopathologic images processing problems. The review showed increasing use of deep learning methods and a constant use of shallow methods. It also showed that the HI processing is an increasing topic of interest. Based on the literature it is also possible to notice a gap between the "old" methods (e.g. hand-crafted feature extractors, shallow classifiers, and image processing) and deep methods. The use of "old" methods may help to solve the lack of data to use the automatic feature extraction from deep methods. A merge of prior knowledge and automatic learning may be profitable for this type of classification.

We proposed the use of patching procedure as a data augmentation method as other works. Our proposal differs from the others in the point that we intend to use a mechanism to create better patches. The first attempt performed well, but with little margin of improvement. It was based on transfer learning from one dataset of colorectal histopathology images to other of breast cancer. 

%The other approach is in a testing phase and involves the use of activation maps of CNNs to identify important regions for classification.

%We consider yet bring knowledge from other feature extractors and pre-processing techniques used in the past to aggregate information and help deep methods to perform well with a small amount of data. If we succeed on that, we may be able to explore this technique to other problems with similar characteristics.



%% References


\section*{References}

\bibliographystyle{model1-num-names}
\bibliography{sample.bib}






% \end{thebibliography}

\end{document}